\documentclass{article}

\usepackage{microtype}
\usepackage{graphicx}
\usepackage{subcaption}
\usepackage{booktabs} 

\usepackage{hyperref}

\usepackage[accepted]{icml2024}

\usepackage{amsmath}
\usepackage{amssymb}
\usepackage{mathtools}
\usepackage{amsthm}

\newcommand{\x}{\boldsymbol{x}}
\newcommand{\xu}{\boldsymbol{u}}

\def\algo{{\textsc{FineSSL}}}

\usepackage[capitalize,noabbrev]{cleveref}
\crefname{section}{Sec.}{Secs.}
\Crefname{section}{Section}{Sections}
\Crefname{table}{Table}{Tables}
\crefname{table}{Tab.}{Tabs.}

\theoremstyle{plain}
\newtheorem{theorem}{Theorem}[section]

\theoremstyle{definition}
\newtheorem{definition}[theorem]{Definition}

\theoremstyle{remark}

\usepackage[textsize=tiny]{todonotes}

\icmltitlerunning{Erasing the Bias: Fine-Tuning Foundation Models for Semi-Supervised Learning}

\newcommand{\ms}[2]{{#1}{\footnotesize $\,\pm${#2}}}

\usepackage{multirow}
\renewcommand{\algorithmiccomment}[1]{\hfill $\triangleright$ #1}

\begin{document}

\twocolumn[
\icmltitle{Erasing the Bias: Fine-Tuning Foundation Models for Semi-Supervised Learning}



\icmlsetsymbol{equal}{*}

\begin{icmlauthorlist}
\icmlauthor{Kai Gan}{seu}
\icmlauthor{Tong Wei}{seu}
\end{icmlauthorlist}

\icmlaffiliation{seu}{School of Computer Science and Engineering, Southeast University, Nanjing 210096, China}

\icmlcorrespondingauthor{Tong Wei}{weit@seu.edu.cn}

\icmlkeywords{Machine Learning, ICML}

\vskip 0.3in
]



\printAffiliationsAndNotice{}  

\begin{abstract}
Semi-supervised learning (SSL) has witnessed remarkable progress, resulting in the emergence of numerous method variations. However, practitioners often encounter challenges when attempting to deploy these methods due to their subpar performance. In this paper, we present a novel SSL approach named \algo\ that significantly addresses this limitation by adapting pre-trained foundation models. We identify the \textit{aggregated biases} and \textit{cognitive deviation} problems inherent in foundation models, and propose a simple yet effective solution by imposing balanced margin softmax and decoupled label smoothing. Through extensive experiments, we demonstrate that \algo\ sets a new state of the art for SSL on multiple benchmark datasets, reduces the training cost by over six times, and can seamlessly integrate various fine-tuning and modern SSL algorithms. The source code is available at \url{https://github.com/Gank0078/FineSSL}.
\end{abstract}

\section{Introduction}
\label{submission}

Semi-supervised learning (SSL) has emerged as a prominent learning paradigm of machine learning, which aims to train models using a combination of a large amount of unlabeled data and a limited number of labeled samples. Recently, numerous SSL algorithms have been proposed to address this challenge by automatically generating pseudo-labels for unlabeled samples using the model and selecting reliable ones for subsequent training \cite{lee2013pseudo,bachman2014learning,laine2016temporal,sajjadi2016regularization,xie2020self,rizve2021defense,wang2022freematch}. Specifically, these algorithms consider model predictions with confidence higher than a specified threshold as reliable pseudo-labels, which are then used to calculate the unsupervised loss. Modern SSL methods like FixMatch \cite{sohn2020fixmatch} and FlexMatch \cite{zhang2021flexmatch} determine the use of pseudo-labels based on a fixed threshold of 0.95 or dynamic thresholds, employing consistency regularization to enforce consistent predictions between weak and strong augmented views of an image.

\begin{figure}[t]
  \centering
  \begin{subfigure}[b]{0.238\textwidth}
  \centering
  \includegraphics[width=\linewidth]{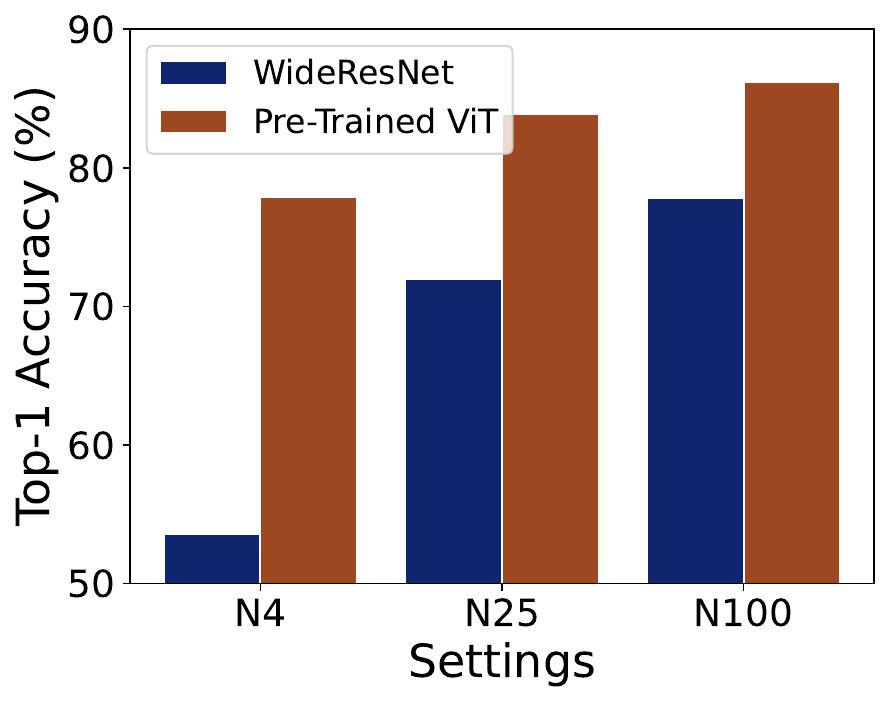}
  \end{subfigure}
 \begin{subfigure}[b]{0.238\textwidth}
  \centering
  \includegraphics[width=\linewidth]{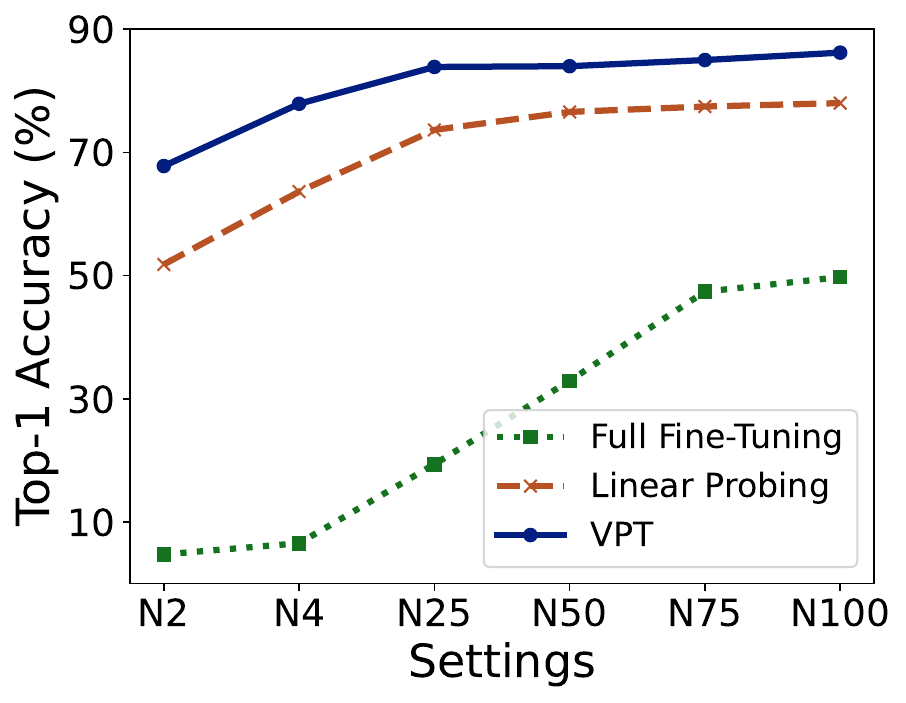}
 \end{subfigure}
  \vskip -0.1in
  \caption{\textit{Left}: Fine-tuning pre-trained ViT significantly outperforms training Wide ResNet starting from scratch. \textit{Right}: VPT improves full fine-tuning and linear probing by a large margin. Experiments are conducted on CIFAR-100 using FixMatch. Throughout the paper, we denote the setting with 4 labeled samples for each class as ``N4'', and other settings are defined accordingly.
  }\label{fig:res_vit_bar}
\end{figure}

While current SSL algorithms have shown promising performance on various tasks, there is still a gap when it comes to deploying them in real-world applications, especially in scenarios with extremely limited labeled data. To address this issue, we propose to develop a more effective SSL approach based on foundation models, \textit{e.g.,} CLIP \cite{radford2021learning}, instead of training models from scratch. These foundation models are pre-trained on large-scale pretext datasets and possess the ability to learn generalizable representations that can be transferred to different tasks. In fact, the adaptation of foundation models has already demonstrated effectiveness in various downstream tasks, including few-shot learning \cite{liu2022few}, long-tail learning \cite{shi2023parameter}, and natural language understanding \cite{yang2022parameter}. However, the potential of foundation models in improving SSL remains unexplored. To motivate our approach, we first find that visual prompt tuning (VPT) \cite{jia2022visual}, a representative parameter-efficient fine-tuning (PEFT) method, is better suited for SSL tasks compared to commonly used full fine-tuning (FFT) and linear probing (LP). Note that PEFT keeps the pre-trained model frozen and only learns a small set of task-specific parameters for adaptation, while FFT updates the entire neural network and LP modifies only the linear classifier. As depicted in \Cref{fig:res_vit_bar}, fine-tuning pre-trained vision Transformer yields significantly better results than training Wide ResNet from scratch \cite{zagoruyko2016wide}, and VPT outperforms other fine-tuning methods across all SSL settings.

\begin{figure}[t]
  \centering
  \begin{subfigure}[b]{0.238\textwidth}
  \centering
  \includegraphics[height=3.15cm,width=\linewidth]{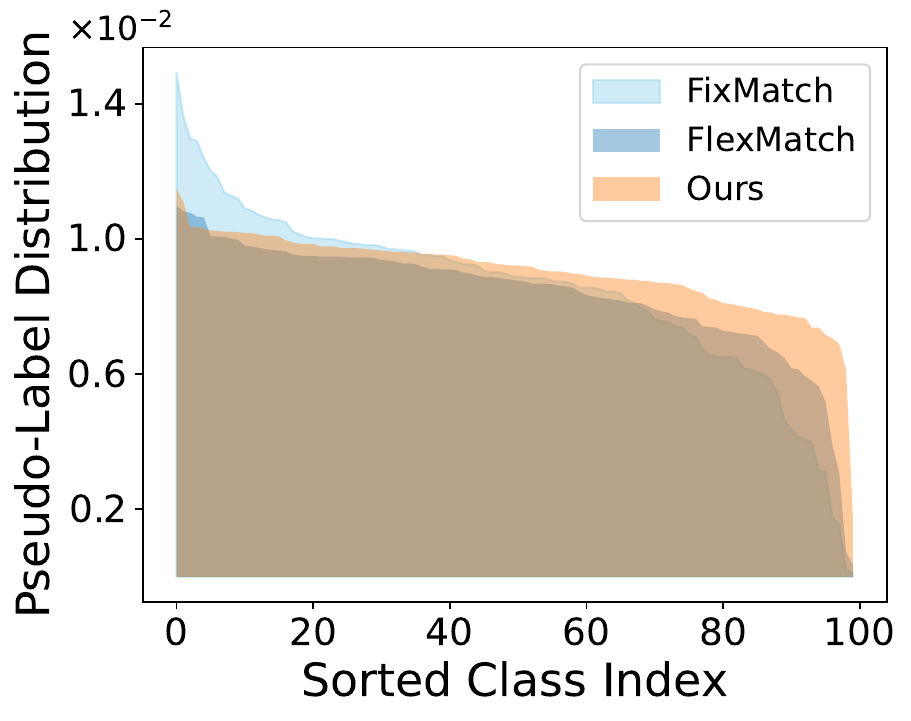}
  \end{subfigure}
 \begin{subfigure}[b]{0.238\textwidth}
  \centering
  \includegraphics[height=3.1cm,width=\linewidth]{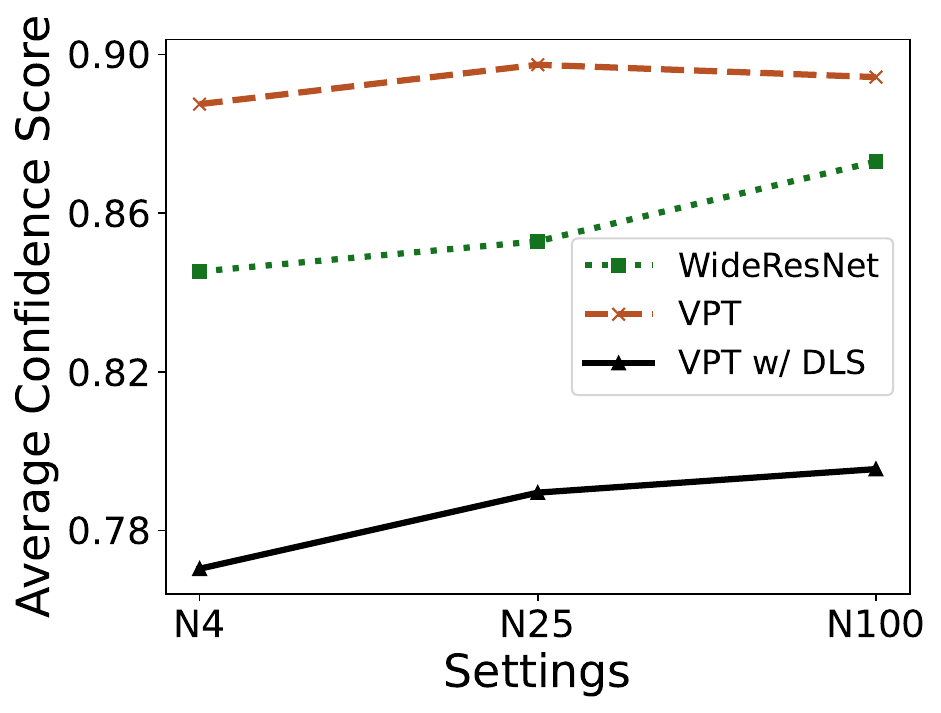}
 \end{subfigure}
  \vskip -0.1in
  \caption{\textit{Left}: The distribution of pseudo-labels for unlabeled data. Classes are sorted by the frequencies of pseudo-labels for each class. \textit{Right}: The average confidence across settings with different numbers of labeled samples per class based on FixMatch. ``DLS'' denotes the decoupled label smoothing proposed in \Cref{method:dls}. Experiments are conducted on CIFAR-100.}\label{fig:ulab_area}
\end{figure}

Considering that PEFT helps improve the performance of pre-trained models in downstream tasks, we are intrigued by its potential as a versatile approach for combining mainstream SSL methods and leveraging the full potential of unlabeled data. However, upon investigating this question, we uncover that fine-tuning foundation models is a double-edged sword. While it improves performance over random initialization, it can also have adverse effects on the selection of reliable pseudo-labels during training. We identify two perspectives from which this issue arises.


One of the perspectives we identified is that foundation models can exhibit biases among different groups of classes due to the inherent imbalances in their pre-training datasets \cite{zhu2023generalized}. Consequently, they tend to generate more pseudo-labels for classes that occur more frequently, particularly when employing strategies like VPT, enabling the reservation of more pre-trained representations \cite{yang2022parameter}. This confirmation bias \cite{arazo2020pseudo,wang2021self} becomes more pronounced as the training progresses, leading to the selection of incorrect pseudo-labels with high model predictive confidence. We refer the amplification of the foundation model's inherent imbalances during the training of SSL methods due to conformation bias as \textit{aggregated biases}. We illustrate its impact on the distribution of pseudo-labels generated by FixMatch and FlexMatch in the left panel of \Cref{fig:ulab_area}. The figure clearly shows that FixMatch exhibits a significantly imbalanced distribution of pseudo-labels, while FlexMatch partially mitigates this issue by dynamically adjusting confidence thresholds based on the learning difficulties of different classes. However, there is still significant room for improvement in addressing the challenge of {aggregated biases}.


The second perspective we identified is that the predictive confidence of foundation models is not reliable. More concretely, this can be manifested in holistic and individual aspects. From the holistic aspect, we observe that confidence scores cannot accurately reflect the true learning difficulties of different SSL tasks. The right panel of \Cref{fig:ulab_area} provides an intuitive demonstration of this finding. In this experiment, we compare the average predictive confidence of FixMatch using Wide ResNet and VPT across three different settings representing varying levels of learning difficulties. Overall, the Wide ResNet model tends to become more confident in its predictions as the number of labeled samples in the training data increases. However, VPT tends to be overconfident  \cite{thulasidasan2019mixup,chan2020unlabelled} even when each class only has four labeled samples, \textit{i.e.,} the ``N4'' setting. Moreover, the model's confidence level in the ``N25'' setting surpasses that of the ``N100'' setting, leading to inaccurate estimates of learning difficulties. In addition, from the individual aspect, we find that the confidence score fails to effectively distinguish between correctly labeled pseudo-labels, incorrectly labeled pseudo-labels, and out-of-distribution (OOD) samples. As depicted in \Cref{fig:ood1}, the distribution of the confidence score across various samples exhibits substantial overlap, which impedes the ability to reliably select high-quality pseudo-labels. We term the phenomenon as \textit{cognitive deviation} which makes it challenging to choose confidence thresholds for pseudo-label selection.

In this paper, we propose a novel approach called \algo\ to address the issues of {aggregated biases} and {cognitive deviation} caused by pre-trained foundation models simultaneously. To mitigate the problem of {aggregated biases}, \algo\ incorporates a balanced margin softmax to compensate for the learning of hard classes and encourage the model to generate more balanced pseudo-labels. The class margins are updated adaptively during the training process.
Unlike previous SSL methods \cite{xu2021dash,wang2022freematch,chen2023softmatch}, which manipulate confidence thresholds to achieve class-balanced pseudo-labels, \algo\ takes a different approach to rectify model preferences directly.
To address the issue of {cognitive deviation}, \algo\ introduces the decoupled label smoothing to regularize the learning pace of an auxiliary classifier, which helps to align the model's confidence with the learning difficulties of tasks. Additionally, \algo\ employs a reweighting strategy to utilize all unlabeled data better. In doing so, it prevents the selection of inaccurate pseudo-labels by thresholding unreliable model confidence. As shown in \Cref{fig:ulab_area}, \algo\ significantly improves the baselines in terms of both pseudo-label distribution and average model confidence.
In summary, our key contributions are:
\vspace{-\topsep}
\begin{itemize}
\setlength\itemsep{0.1em}
\item We identify the {aggregated biases} and {cognitive deviation} problems caused by fine-tuning foundation models. We find that PEFT significantly outperforms FFT and LP  in SSL tasks.
\item  We propose the balanced margin softmax to erase aggregated biases, decoupled label smoothing to reduce cognitive deviation, and a reweighting strategy to take full advantage of unlabeled data.
\item We propose a general fine-tuning framework for SSL, which can smoothly incorporate various PEFT and leading SSL methods for improving performance.
\item Extensive experimental analyses show that \algo\ achieves state-of-the-art performance on multiple SSL benchmarks. Moreover, \algo\ obtains significant reductions in the training cost.
\end{itemize}

\section{Related Works}

\textbf{Semi-Supervised Learning.} 
Modern SSL methods typically combine multiple techniques for leveraging unlabeled data, including entropy minimization \cite{grandvalet2004semi}, consistency regularization \cite{sohn2020fixmatch}, distribution alignment \cite{kim2020distribution, wei2023towards}, and contrastive learning \cite{lee2022contrastive, yang2022class}. Most of them involve selecting reliable pseudo-labels during the training process. Initially, FixMatch selects pseudo-labels with confidence higher than a fixed threshold of 0.95. Recently, some works mention that a fixed threshold can be too strict for classes that are hard to learn, and propose to dynamically adjust the threshold as the training progresses. For example, FlexMatch \cite{zhang2021flexmatch} assigns distinct thresholds to each class based on their levels of learning difficulty as well as the overall learning pace. FreeMatch \cite{wang2022freematch} integrates global and local threshold adjustments while incorporating class-fairness regularization, which encourages the model to make more diverse predictions. SoftMatch \cite{chen2023softmatch} achieves a quantity-quality trade-off through unified sample reweighting by employing a soft confidence threshold mechanism. In contrast, this paper encourages the model to explicitly produce class-balanced pseudo-labels rather than adjusting confidence thresholds.

\textbf{SSL based on Foundation Models.} A few prior efforts have been made to leverage pre-trained models to improve the performance of SSL tasks. As a representative, SimCLRv2 \cite{chen2020big} employs a pre-trained ResNet model \cite{he2016deep} using self-supervised learning, and fine-tunes it using the labeled data. \citet{zhou2018semi} delve comprehensively into the performance dynamics of SSL methods originating from pre-trained models, exploring their effectiveness across varying conditions. USB \cite{wang2022usb} introduces pre-trained Transformers \cite{vaswani2017attention} into SSL, which reduces 80\% training steps without hurting the performance. However, USB merely explores the FFT to adapt the pre-trained model, which is inferior to PEFT in SSL tasks. Recently, DebiasPL \cite{wang2022debiased} integrates large-scale multi-modal pre-trained model CLIP \cite{radford2021learning} into FixMatch by pseudo-labeling the discarded unlabeled instances with CLIP and effectively adjusts the distribution of pseudo-labels by logit adjustment \cite{menon2020long} according to the distribution of pseudo-labels. However, how to effectively fine-tune the foundation models in SSL tasks remains underexplored.

\section{Preliminaries}

\subsection{Background of Semi-Supervised Learning}

First of all, we describe the problem setup and define notations frequently used throughout the paper. The goal of SSL is to learn a $C$-class classifier using a large number of unlabeled samples coupled with a few labeled ones. Let $\mathcal{D}^{l}=\{(\x_i,y_i)\}^{N}_{i=1}$ denote the labeled dataset of size $N$ and $\mathcal{D}^{u}=\{\xu_j\}^{M}_{j=1}$ denote the unlabeled dataset of size $M$. Here, $\x_i, \xu_j \in \mathbb{R}^d$ are the $d$-dimensional feature vectors. For each labeled sample $\x_i$, we have access to its ground-truth class label $y_i \in \{0, 1\}^{C}$. Typically, we have $N \ll M$ in SSL tasks. Then, our goal is to learn a model $f(\x)$ parameterized by $\theta$ utilizing $\mathcal{D}^{l}$ and $\mathcal{D}^{u}$. 

Modern SSL approaches usually have two components in their learning objectives, \textit{i.e.,} the supervised and unsupervised losses \cite{sohn2020fixmatch}. Specifically, the standard cross-entropy $\mathcal{H}$ is utilized to optimize the supervised loss $\ell_s$ using labeled dataset:
\begin{equation}
    \ell_s = \frac{1}{B} \sum_{i=1}^B \mathcal{H}(y_i, p(y \mid \x_i))
    \label{eq:ce_lab}
\end{equation}
where $p(y \mid \x_i) = \mathrm{Softmax}(f(\x_i; \theta))$ denotes the posterior probability of $\x_i$ being classified into class $y$, and $B$ represents the batch size for labeled data.

Regarding the unsupervised loss, the majority of existing methods rely on pseudo-labeling \cite{lee2013pseudo, berthelot2019mixmatch, sohn2020fixmatch, zhang2021flexmatch, wang2022freematch, chen2023softmatch}. To ensure the quality of generated pseudo-labels, these methods commonly employ a thresholding mechanism to choose pseudo-labels with higher confidence and impose consistency regularization on them. The widely used consistency regularizer proposed in FixMatch \cite{sohn2020fixmatch} is defined as follows:
\begin{equation}
    \ell_u = \frac{1}{\mu B} \sum_{j=1}^{\mu B} \mathbb{I}\left(\max \left(\boldsymbol{q}_j\right)  \ge \tau \right) \mathcal{H}(\widehat{q}_j, p(y \mid \Omega(\xu_j)))
    \label{eq:ce_ulab}
\end{equation}
where $\mu$ is a hyperparameter used to control the relative batch size ratio between labeled and unlabeled data, $\boldsymbol{q}_j = p(y \mid \omega(\xu_j))$, and $\widehat{q}_j=\arg\max_k (q_{jk})$ is the pseudo-label of $\xu_j$. $\tau$ is a threshold to select reliable pseudo-labels for calculating the loss and $\mathbb{I}(\cdot)$ is the indicator function. Since applying the consistency regularizer needs two different views of images, we use $\omega(\cdot)$ and $\Omega(\cdot)$ to represent the weak and strong augmented versions of $\xu_j$, respectively.

\subsection{Fine-tuning Foundation Models}

The prevailing trend in modern deep learning is the fine-tuning of large-scale pre-trained vision Transformers \cite{dosovitskiy2020image,radford2021learning,jia2021scaling} by leveraging rich semantic information to augment the performance of downstream tasks. Common fine-tuning strategies are FFT and LP, wherein the former involves adjusting the parameters of the entire pre-trained model, while the latter entails solely fine-tuning the final linear classifier of the neural networks. As depicted in \Cref{fig:res_vit_bar}, FFT achieves inferior performance because the limited supervisory information available in SSL is not enough to guide the training of a big model. In comparison, LP performs reasonably well by freezing the pre-trained parameters except the linear classifier. However, since there exists a gap between pre-training datasets and our downstream SSL dataset, the learned representations can be suboptimal, which hinders the improvement of performance.
Therefore, we propose to employ PEFT to effectively adapt the representations and the classifier. Although many PEFT methods can be utilized, we use VPT \cite{jia2022visual} as the default choice without loss of generality. The rationale of VPT is simple. The Transformers divide an input image into multiple patches of the same size and we denote $[\boldsymbol{c}_l; \boldsymbol{E}_l]$ as the input for $l$-th layer of ViT for simplicity. Here, $\boldsymbol{c}_l$ is the extra learnable token for classification and $\boldsymbol{E}_l$ denotes the collection of image patch embeddings. VPT prepends learnable prompts $\boldsymbol{P}_l$ at each layer to extend $[\boldsymbol{c}_l; \boldsymbol{E}_l]$ to $[\boldsymbol{c}_l; \boldsymbol{P}_l; \boldsymbol{E}_l]$. VPT has two variations: 1) VPT-shallow, where prompts are appended solely at the first layer; 2) VPT-deep, characterized by the extension of prompts to all layers. In this paper, we focus on VPT-deep for its stronger generalization capability. More analyses on different PEFT methods can be found in \Cref{sec:deeper_look}. To simplify the mathematical notation, we denote the parameters of the pre-trained foundation model as $\Theta$ and the parameters of the PEFT modules, including the final linear classifier, as $\theta$.

\section{The Proposed \algo\ Framework}

In this section, we present the \algo\ framework in detail by introducing two of its core components, \textit{i.e.,} balanced margin softmax and decoupled label smoothing.

\subsection{Balanced Margin Softmax}
\label{sec:debiasing}

Our empirical studies in \Cref{fig:ulab_area} highlight that employing dynamic thresholds for classes based on learning difficulties may not sufficiently mitigate {aggregated biases}. Instead of focusing on pseudo-label selection, we propose to explicitly intervene in the model training process. Drawing inspiration from previous works in class-imbalanced learning \cite{cao2019learning,menon2020long,ye2023bridging}, we devise a balanced margin softmax loss with enforced dynamic class margins into the standard cross-entropy. For each training sample $(\x, y)$, we define:
\begin{equation}
    \mathcal{H}_{m}(y, \boldsymbol{z}) = - \log 
    \frac{e^{z_y}}
    {e^{z_y} + \sum_{k \neq y} e^{z_k + \alpha_t \Delta^t_{y}}}
    \label{eq:margin}
\end{equation}
where $k \in \{1, \ldots, C\}$, $\boldsymbol{z} = f(\x; \{\Theta, \theta\})$ denotes the output logits, and $\Delta_{k}^t$ is the margin for class $k$ at $t$-th training iteration. $\alpha_t$ is a scaling hyperparameter of class margins which is adaptively updated during training. 
Instead of using a constant margin for all classes, the incorporation of class-specific margins helps improve inter-class discrimination and makes it more flexible to control the learning pace of classes.
In \Cref{eq:margin}, we impose a large positive margin $\Delta_{y}^t$ when the model exhibits weak preference for predicting class $y$. This encourages the model to be more confident in its predictions in the next training iteration, gradually erasing the inherent bias of the foundation model and aggregated biases.

Next, we present a simple approach to automatically assign class margins based on model dynamics. We start by introducing the concept of learning pace for each class, which describes the model's capability of producing accurate pseudo-labels.
\begin{definition}{\textit{Class Learning Pace.}}
The learning pace of the $k$-th class $\sigma_{t}(k)$ at timestamp $t$ is defined as the number of unlabeled samples predicted as class $k$ with confidence surpassing a constant $\zeta$, \textit{i.e.,}
    \begin{equation}
\resizebox{0.91\hsize}{!}{
    $\sigma_{t}(k) = \sum_{j=1}^M \mathbb{I}\left(\max \left(\boldsymbol{q}_j\right)  \ge \zeta \right) \cdot \mathbb{I}\left(\arg\max \left(\boldsymbol{q}_j\right)  = k \right)
    \label{eq:difficulty}$.
}
\end{equation}
\end{definition}
The learning pace effectively reflects the model's abilities in each class, with classes having fewer predicted pseudo-labels above the threshold indicating slower learning paces, and vice versa. A similar concept of learning pace is explored in FlexMatch \cite{zhang2021flexmatch}, where confidence thresholds for classes are assigned based on a similar principle. In our implementation, we fix the threshold $\zeta=0.7$ for simplicity, and the learning paces can be calculated by counting the number of confident pseudo-labels for each class. To normalize the learning paces across all classes, we define $\beta_t{(y)} = \frac{\sigma_{t}(y)}{\max_{k} \sigma_{t}(k)}$ and let $\Delta^{t}_{y} = 1 - \beta_t{(y)}$. In doing so, the model pays more attention to classes of slow learning paces. 

Additionally, we use the parameter $\alpha_t$ in \Cref{eq:margin} to control the magnitude of the margin $\Delta^{t}_{y}$. Ideally, $\alpha_t$ is supposed to be $0$ if the learning pace of all classes is approximately equal and the balanced margin softmax reduces to the standard cross-entropy. Conversely, a large $\alpha_t$ is desired to enhance the effect of the balanced margin softmax. Given a base scaling parameter $\alpha$, we propose to dynamically adjust its value at the $t$-th iteration based on learning paces via:
\begin{equation}
    \alpha_{t} =  (\max_{k} (\beta_t(k)) - \min_{k} (\beta_t(k)))  \alpha .
    \label{eq:alpha}
\end{equation}
Once obtaining $\alpha_{t}$ and $\Delta^{t}_{y}$, we can improve the balancedness of pseudo-labels and reduce aggregated biases by minimizing \Cref{eq:margin}. While the learning paces are estimated based on unlabeled data, they effectively capture the learning dynamics of the model. By imposing balanced margins over the labeled data, we can further improve the model's generalization capabilities. To achieve this, we replace the standard cross-entropy with the balanced margin softmax for both the supervised loss and consistency regularization.

\subsection{Decoupled Label Smoothing}
\label{method:dls}

While the balanced margin softmax alleviates the imbalance issue caused by pre-trained foundation models, the presence of {cognitive deviation} in the training process can still lead to suboptimal performance in selecting reliable pseudo-labels based on confidence thresholds. To address this, we propose a simple yet effective approach called decoupled label smoothing to overcome the mismatched model confidence and task difficulties. Label smoothing is a well-known technique that has been successfully applied across a range of tasks to regularize model training and mitigate overconfidence issues \cite{muller2019does,zhong2021improving}. In a nutshell, label smoothing regularizes the model training by softening the ground-truth labels. However, directly applying label smoothing to the learning objective can negatively impact representation learning, as demonstrated in our ablation studies. To overcome this limitation, we introduce an auxiliary classifier $f_{aux}(\x)$, which is a single fully connected layer appended to the feature extractor. To distinguish $f_{aux}(\x)$ from $f(\x)$, we denote its output logits using $f_{aux}((\x); \{\Theta, \theta^\prime\})$, where $\theta^\prime$ contains parameters for PEFT modules and the auxiliary classifier. This auxiliary classifier is learned by minimizing the supervised loss $\ell_s$ and the soft version of consistency regularization:
\begin{equation}
    \ell^{ls}_u = \frac{1}{\mu B} \sum_{j=1}^{\mu B} \mathcal{H}_{m}(\widetilde{\boldsymbol{q}}_j, f_{aux}(\Omega(\xu_j); \{\Theta, \theta^\prime\}))
    \label{eq:aux_ls}
\end{equation}
where $\widetilde{\boldsymbol{q}}_j$ denotes the continuous pseudo-label vector for $\xu_j$, $\widetilde{{q}}_{jk}$ equals $(1 - \lambda) + \lambda / C$ if $k=\widehat{q}_j$, otherwise $\lambda / C$. $\lambda \in (0, 1)$ is the label smoothing parameter.

It is crucial to note that the auxiliary classifier is intentionally decoupled from the main branch of neural networks. This ensures that the gradients of the auxiliary classifier's loss function do not propagate to the earlier layers during backpropagation. As a result, this prevents label smoothing from damaging the representation learning and allows it to focus solely on generating discriminative weights between correct and incorrect pseudo-labels.

\begin{algorithm}[tb]
   \caption{The Proposed \algo}
   \label{alg:ours}
\begin{algorithmic}[1]
   \STATE {\bfseries Input:} labeled dataset $\mathcal{D}^{l}$ and unlabeled dataset $\mathcal{D}^{u}$, pre-trained foundation model, hyperparameters $\{\zeta, \alpha, \lambda\}$, number of iterations $T$
   \FOR{$t=1$ {\bfseries to} $T$}
   \STATE Update learning pace $\sigma_{t}(y)$ by \Cref{eq:difficulty}
   \IF{$\max_{y} (\sigma_{t}(y)) > 0$}
   \STATE $\beta_t{(y)} = \frac{\sigma_{t}(y)}{\max_{k} \sigma_{t}(k)}$ \hfill \COMMENT{Normalization}
   \STATE $\Delta^{t}_{y} = 1 - \beta_t{(y)}$  \COMMENT{Obtain class-specific margin}
   \ELSE
   \STATE $\Delta^{t}_y = 1$
   \ENDIF
   \STATE Compute scaling parameter $\alpha_{t}$ by \Cref{eq:alpha}
   \STATE Compute unlabeled sample weights $\psi(\xu)$
   \STATE Compute $\ell_s^m$ and $\widetilde{\ell}^m_u$ for the main branch
    \STATE Compute $\ell_s$ and $\ell^{ls}_u$ for the auxiliary classifier
   \STATE Compute the total loss $\mathcal{L}$ by \Cref{eq:total_loss}
   \STATE Update $\theta$ based on $\nabla\mathcal{L}$ using SGD
   \ENDFOR
\end{algorithmic}
\end{algorithm}

Next, we aim to resolve another substantial disadvantage of pseudo-label selection, which merely utilizes unlabeled data of high confidence, and low-confidence data are discarded. Obtaining discriminative predictive probabilities from the auxiliary classifier, we take full advantage of unlabeled data to optimize the model by incorporating sample reweighting into consistency regularization. Specifically, we define the weighted consistency regularizer as follows:
\begin{equation}
    \widetilde{\ell}^m_u = \frac{1}{\mu B} \sum_{j=1}^{\mu B} \psi(\xu_j) \mathcal{H}_{m}(\widehat{q}_j, f(\Omega(\xu_j); \{\Theta, \theta\}))
    \label{eq:ulab_weight}
\end{equation}
where $\psi(\xu_j) = \gamma \cdot \max (p_{aux}(y \mid \omega(\xu_j))$ represents the importance weight for the $j$-th unlabeled sample, and $\gamma$ is a tunable hyperparameter. The output probabilities $p_{aux}(y \mid \omega(\xu_j)) = \mathrm{Softmax}(f_{aux}(\omega(\xu_j); \{\Theta, \theta^\prime\}))$. Notice that the auxiliary classifier is solely employed for generating sample weights, which facilitates the learning of the main branch. In return, better representations learned by the main branch improve the training of the auxiliary classifier. We analyze model confidence used for sample reweighting in \Cref{app:cd}.

To sum up, the overall loss for the proposed method comprises two parts, \textit{i.e.,} losses for the main branch and auxiliary classifier, which are formulated as follows:
\begin{equation}
\resizebox{0.6\hsize}{!}{
    $\mathcal{L} = \underbrace{\ell^m_s + \widetilde{\ell}^m_u}_{\mathsf{main \ branch}} + \underbrace{\ell_s + \ell^{ls}_u}_{\mathsf{auxiliary \ classifier}}$.
}
    \label{eq:total_loss}
\end{equation}
Here, we denote $\ell^m_s$ as the balanced margin softmax loss over labeled data.
We present the pseudo-code of the proposed \algo\ algorithm in \Cref{alg:ours}.

\begin{table*}[t]
\caption{Performance comparisons on CIFAR-10, CIFAR-100, and FOOD-101 datasets: mean $\pm$ std of accuracy over 3 trials are reported. The best performance is highlighted in \textbf{bold} and the second-best performance is \underline{underlined}.}
\label{tab:main}
\begin{center}
\begin{sc}
\resizebox{\textwidth}{!}{
\begin{tabular}{@{}lccccccccc@{}}
\toprule
 & \multicolumn{3}{c}{CIFAR-10} & \multicolumn{3}{c}{CIFAR-100} & \multicolumn{3}{c}{FOOD-101} \\
\cmidrule(rl){2-4} \cmidrule(rl){5-7} \cmidrule(l){8-10}

Settings & N1        & N2        & N4       & N4        & N25       & N100     & N2        & N4       & N10      \\ \midrule
Supervised      & \ms{68.41}{6.60}    & \ms{78.54}{4.94}    & \ms{86.94}{4.17}   & \ms{64.18}{0.33}    & \ms{80.23}{0.14}    & \ms{84.45}{0.29}    & \ms{61.42}{0.21}    & \ms{73.74}{0.22}   & \ms{82.10}{0.13}   \\ \midrule
\begin{tabular}[c]{@{}l@{}}PL\\ FixMatch\\ FlexMatch\\ FreeMatch\\ SoftMatch\\ DebiasPL\end{tabular} &
  \begin{tabular}[c]{@{}c@{}}\ms{48.45}{2.99}\\ \ms{60.27}{1.89}\\ \ms{61.71}{15.07}\\ \ms{60.41}{1.78}\\ \ms{59.74}{11.44}\\ \ms{\underline{78.30}}{16.85}\end{tabular} &
  \begin{tabular}[c]{@{}c@{}}\ms{74.33}{10.97}\\ \ms{87.31}{0.10}\\ \ms{92.71}{7.31}\\ \ms{87.04}{0.06}\\ \ms{90.72}{5.28}\\ \ms{\underline{96.83}}{0.14}\end{tabular} &
  \begin{tabular}[c]{@{}c@{}}\ms{92.80}{0.38}\\ \ms{\textbf{97.43}}{0.14}\\ \ms{\underline{97.41}}{0.10}\\ \ms{97.26}{0.14}\\ \ms{97.35}{0.06}\\ \ms{97.39}{0.06}\end{tabular} &
  \begin{tabular}[c]{@{}c@{}}\ms{66.17}{0.30}\\ \ms{77.10}{1.04}\\ \ms{79.19}{0.23}\\ \ms{76.19}{0.82}\\ \ms{75.94}{0.46}\\ \ms{\underline{79.57}}{0.32}\end{tabular} &
  \begin{tabular}[c]{@{}c@{}}\ms{81.45}{0.30}\\ \ms{\underline{84.05}}{0.16}\\ \ms{83.77}{0.13}\\ \ms{83.60}{0.16}\\ \ms{83.85}{0.06}\\ \ms{84.01}{0.07}\end{tabular} &
  \begin{tabular}[c]{@{}c@{}}\ms{84.65}{0.20}\\ \ms{86.17}{0.03}\\ \ms{86.23}{0.25}\\ \ms{\underline{86.34}}{0.30}\\ \ms{86.31}{0.07}\\ \ms{86.16}{0.13}\end{tabular} &
  \begin{tabular}[c]{@{}c@{}}\ms{58.22}{1.18}\\ \ms{74.92}{1.09}\\ \ms{80.96}{1.68}\\ \ms{74.31}{4.56}\\ \ms{75.82}{5.24}\\ \ms{\underline{86.51}}{0.74}\end{tabular} &
  \begin{tabular}[c]{@{}c@{}}\ms{77.57}{0.93}\\ \ms{84.93}{0.70}\\ \ms{88.51}{0.13}\\ \ms{84.69}{1.11}\\ \ms{85.05}{1.05}\\ \ms{\underline{88.60}}{0.34}\end{tabular} &
  \begin{tabular}[c]{@{}c@{}}\ms{84.62}{0.12}\\ \ms{88.72}{0.02}\\ \ms{89.24}{0.11}\\ \ms{88.97}{0.24}\\ \ms{89.01}{0.08}\\ \ms{\underline{89.32}}{0.17}\end{tabular} \\ \midrule
\algo            & \ms{\textbf{96.15}}{0.13}    & \ms{\textbf{96.87}}{0.10}    & \ms{97.39}{0.05}   & \ms{\textbf{80.44}}{0.24}    & \ms{\textbf{84.51}}{0.01}    & \ms{\textbf{86.66}}{0.18}    & \ms{\textbf{87.04}}{0.43}    & \ms{\textbf{89.28}}{0.14}   & \ms{\textbf{89.69}}{0.15}   \\ \bottomrule
\end{tabular}
}
\end{sc}
\end{center}
\end{table*}

\section{Experiments}

\subsection{Experiment Setup}

\noindent
\textbf{Datasets and Evaluation Metrics.} We conduct extensive experiments on five publicly available datasets to evaluate the performance of \algo. The datasets are CIFAR-10, CIFAR-100 \cite{krizhevsky2009learning}, FOOD-101 \cite{bossard2014food}, Semi-Aves \cite{su2021semi}, and ImageNet \cite{deng2009imagenet}. For each dataset, we consider different numbers of labeled samples to cover various SSL scenarios.  Following previous works \cite{wang2022freematch, chen2023softmatch}, we select \{1, 2, 4\} labeled samples per class for CIFAR-10, \{4, 25, 100\} labeled samples per class for CIFAR-100, and \{2, 4, 10\} labeled samples per class for FOOD-101. For Semi-Aves, we consider two scenarios of unlabeled data, \textit{i.e.,} unlabeled data consists of either in-distribution samples or mixed in-distribution and OOD samples. Finally, for ImageNet, we consider the settings where 1\% and 10\% training samples are labeled following SimCLR \cite{chen2020simple}.
In our experiments, we compare \algo\ with existing state-of-the-art SSL methods, including Pseudo-Label (PL) \cite{lee2013pseudo}, FixMatch \cite{sohn2020fixmatch}, FlexMatch \cite{zhang2021flexmatch}, FreeMatch \cite{wang2022freematch}, SoftMatch \cite{chen2023softmatch}, and DebiasPL \cite{wang2022debiased}. All competing methods are implemented in our proposed framework. The classification performance is evaluated using the Top-1 accuracy on the test dataset.

\noindent
\textbf{Implementation details.}
We use the publicly available CLIP model\footnote{\url{https://github.com/mlfoundations/open\_clip}} as our
backbone. Model parameters of CLIP are all frozen, and only the image encoder is used during training and inference. In our experiments, we use VPT \cite{jia2022visual} by default to fine-tune the CLIP model due to its effectiveness and efficiency. The length of learnable prompts is set to 50. We also provide the results for more PEFT methods in \Cref{sec:deeper_look}. We employ the Stochastic Gradient Descent (SGD) optimizer with a learning rate of 0.03, utilizing a batch size of 32 alongside a weight decay set at $5 \times 10^{-4}$, and a momentum factor of 0.9. We fine-tune the model for 30 epochs, with each epoch comprising 500 steps. It is noteworthy that all competing methods share the same setting of those hyperparameters. All experiments are conducted in PyTorch with a single NVIDIA RTX 3090 24GB GPU. More details can be found in \Cref{app:implementation}.

\subsection{Main Results}
\noindent
\textbf{Results on standard datasets.}
\Cref{tab:main} reports the test accuracy of competing methods over CIFAR-10, CIFAR-100, and FOOD-101 datasets. Overall, \algo\ achieves superior performance, substantially outperforming the other seven methods in most datasets and settings. \algo\ presents a relatively low performance in N4 setting on CIFAR-10. This is partly because the task is simple such that several methods achieve comparably high performance \cite{wang2022debiased}. Notably, \algo\ outperforms competing methods by nearly 18\% in accuracy with only $1$ labeled sample per class on CIFAR-10, showing its robust performance in scenarios where labels are extremely scarce. The weak performance of Pseudo-Labeling and FixMatch in low-label regimes can be attributed to the biases and overconfidence issues of the model. By dynamically adjusting the confidence threshold, FlexMatch and FreeMatch achieve better results. DebiasPL addresses the class-imbalanced pseudo-labels and achieves comparable results with \algo\ in two settings on CIFAR-10. However, in other cases, \algo\ surpasses DebiasPL by an average of 3\% in accuracy, showing that \algo\ can perform much more accurately across different settings.

\vskip -0.15in
\begin{table}[ht]
\caption{Top-1 accuracy for Semi-Aves. $\mathcal{D}^{u}_{in}$ denotes that unlabeled data are sampled from the same classes as labeled data, and $\mathcal{D}^{u}_{out}$ contains OOD samples.}
\label{tab:semiaves}
\begin{center}
\begin{sc}
\begin{tabular}{@{}lcc@{}}
\toprule
Settings & $\mathcal{D}^{u} = \mathcal{D}^{u}_{in}$       & $\mathcal{D}^{u} = \mathcal{D}^{u}_{in} \cup \mathcal{D}^{u}_{out}$      \\ \midrule
\begin{tabular}[c]{@{}l@{}}FixMatch\\ FlexMatch\\ FreeMatch\\ SoftMatch\\ DebiasPL\end{tabular} &
  \begin{tabular}[c]{@{}c@{}}\ms{65.52}{0.30}\\ \ms{64.60}{0.25}\\ \ms{64.57}{0.23}\\ \ms{64.75}{0.27}\\ \ms{\underline{66.82}}{0.35}\end{tabular} &
  \begin{tabular}[c]{@{}c@{}}\ms{60.15}{0.16}\\ \ms{57.96}{0.11}\\ \ms{58.01}{0.38}\\ \ms{57.79}{0.52}\\ \ms{\underline{60.85}}{0.37}\end{tabular} \\ \midrule
\algo     & \ms{\textbf{67.25}}{0.11}       & \ms{\textbf{61.12}}{0.07}           \\ \bottomrule
\end{tabular}
\end{sc}
\end{center}
\end{table}

\begin{table}[t]
\caption{Top-1 accuracy for ImageNet. We consider two settings, distinguished by 1\% and 10\% labeled data.}
\label{tab:imagenet}
\begin{center}
\begin{sc}
\begin{tabular}{@{}lcc@{}}
\toprule
Settings & 1\% labeled  & 10\% labeled  \\ \midrule
\begin{tabular}[c]{@{}l@{}}FixMatch\\ FlexMatch\\ FreeMatch\\ SoftMatch\\ DebiasPL\end{tabular} &
  \begin{tabular}[c]{@{}c@{}}73.19\\ 73.21\\ 72.94\\ 72.92\\ \underline{73.58}\end{tabular} &
  \begin{tabular}[c]{@{}c@{}}\underline{78.77}\\ 78.71\\ 78.49\\ 78.56\\ 78.72\end{tabular} \\ \midrule
\algo     & \textbf{74.22}        & \textbf{79.21}         \\ \bottomrule
\end{tabular}
\end{sc}
\end{center}
\vskip -0.2in
\end{table}

\begin{figure*}[!h]
\begin{center}
   \begin{subfigure}[b]{0.24\textwidth}
   \centering
   \includegraphics[width=\linewidth,height=1.25in]{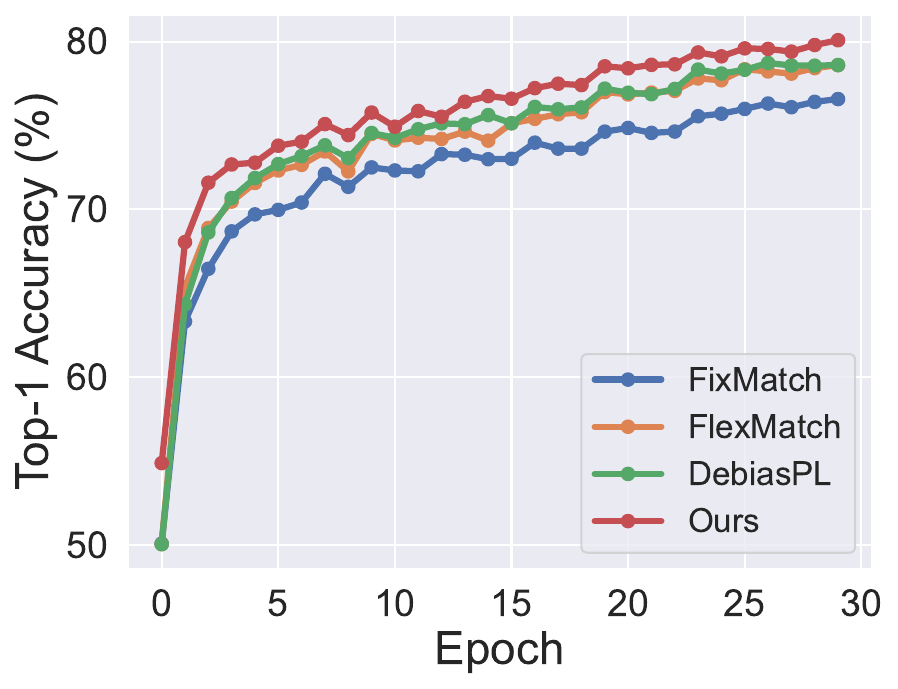}
   \vskip -0.06in

   \caption{}
   \label{fig:ulab_acc}
   \end{subfigure}
   \begin{subfigure}[b]{0.24\textwidth}
   \centering
   \includegraphics[width=\linewidth,height=1.25in]{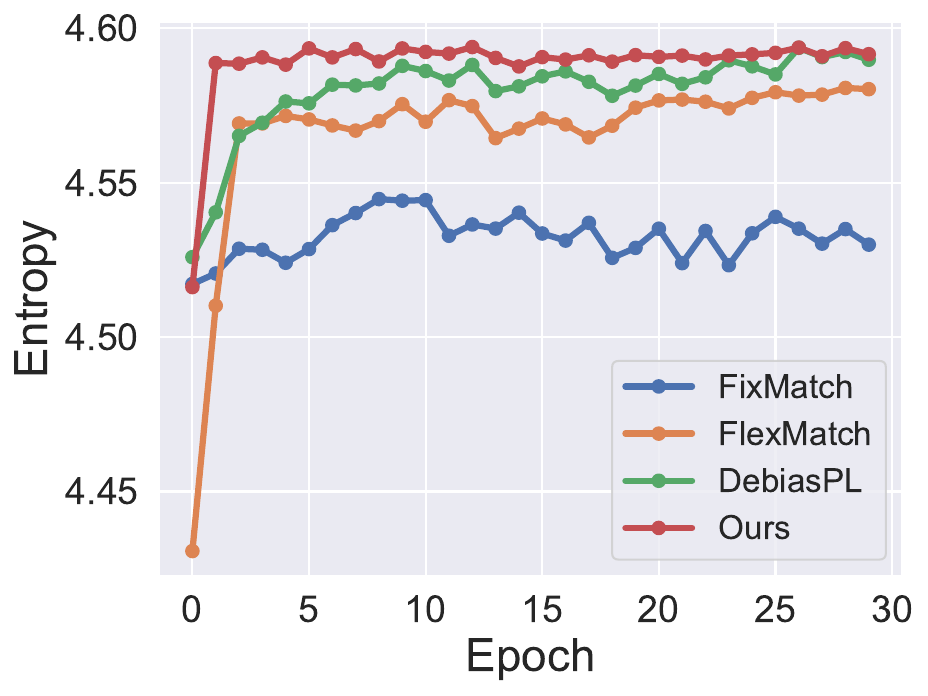}
   \vskip -0.06in
   \caption{}
   \label{fig:ulab_ent}
   \end{subfigure}
   \begin{subfigure}[b]{0.24\textwidth}
   \centering
   \includegraphics[width=\linewidth,height=1.25in]{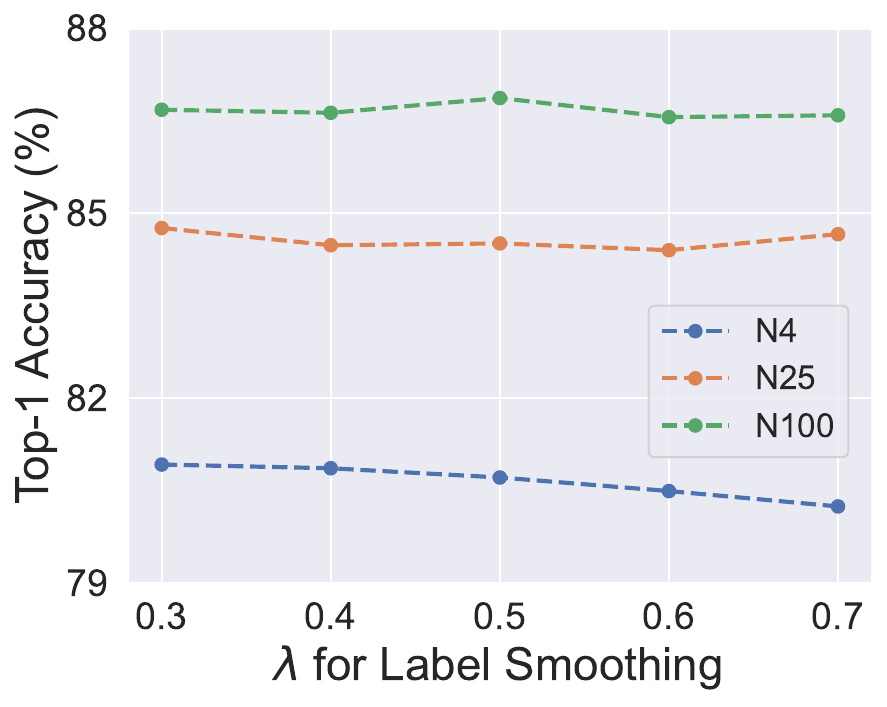}
   \vskip -0.06in

   \caption{}
   \label{fig:sen_lsm}
   \end{subfigure}
   \begin{subfigure}[b]{0.24\textwidth}
   \centering
   \includegraphics[width=\linewidth,height=1.25in]{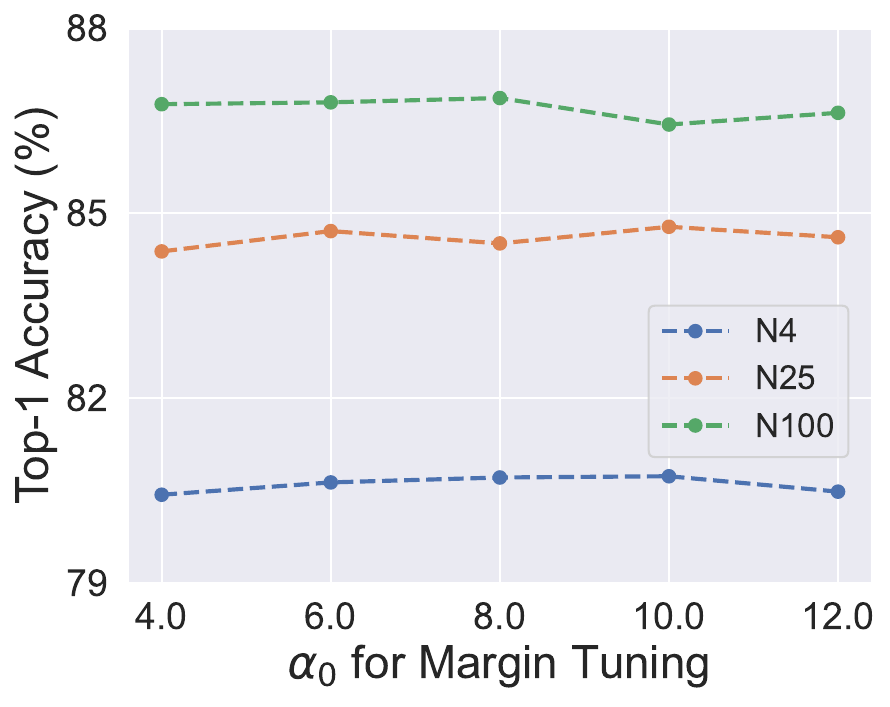}
   \vskip -0.06in

   \caption{}
   \label{fig:sen_alpha}
   \end{subfigure}
  \vskip -0.15in
\caption{(\labelcref{fig:ulab_acc,fig:ulab_ent}): The accuracy and entropy of pseudo-labels for FixMatch, FlexMatch, DebiasPL, and \algo\ on CIFAR100 with 4 labeled data per class. (\labelcref{fig:sen_lsm,fig:sen_alpha}): The sensitivity of $\lambda$ and $\alpha_0$ under various settings on CIFAR-100.}
\label{fig:deeper_look}
\end{center}
\vskip -0.15in
\end{figure*}


\vskip 0.1in
\noindent
\textbf{Results on complex datasets.}
To evaluate the generalization ability of our model, we further examine \algo\ on two more challenging datasets, \textit{i.e.,} Semi-Aves and ImageNet.
\Cref{tab:semiaves} and \Cref{tab:imagenet} show the results for \algo\ and five competeing methods. Semi-Aves \cite{su2021semi} is a dataset of hundreds of bird species. Data exhibits considerable class imbalance, and the unlabeled data contains images from novel classes, which makes the task particularly challenging. We consider both settings where the unlabeled data comes from the same classes with labeled data or contains new classes. FixMatch achieves much better performance than its three sophisticated counterparts, especially when OOD data is present in the unlabeled data. This is attributed to the selection of more OOD samples by decreasing the threshold during training. Our method \algo\ outperforms FixMatch and DebiasPL by 1.4\% and 0.35\% in accuracy on average.
For the ImageNet dataset, we randomly choose 1\% and 10\% images as labeled data following SimCLR \cite{chen2020simple}. Overall, all five competing methods achieve similar performance. By employing \algo, we observe a 0.57\% relative performance increase over DebiasPL. This demonstrates the superior performance of \algo\ in large-scale datasets.

\subsection{A Deeper Look into \algo}
\label{sec:deeper_look}

In this subsection, we conduct qualitative analyses of \algo\ in comparison to other SSL methods. Additionally, we examine the sensitivity of hyperparameters to show the robustness of our method. The comparative evaluations with alternative PEFT strategies and the analysis of training costs are also included.

\noindent
\textbf{Quality of Pseudo-Labels.} From \Cref{fig:ulab_acc}, it is noteworthy that the pseudo-labels generated by \algo\ consistently exhibit higher accuracy during training compared to other methods, and it is obvious that the accuracy of pseudo-labels plays a pivotal role in enhancing the performance of SSL tasks \cite{chen2023softmatch}. In addition, \algo\ shows higher entropy in the pseudo-label distribution during training, indicating that our method effectively mitigates {aggregated biases} and achieves a more balanced distribution of pseudo-labels.

\noindent
\textbf{Hyperparameter Sensitivity.} Despite the commendable performance achieved by our method, the inclusion of some significant parameters prompts us to question whether the outstanding results are attributed to parameter tuning. Therefore, we are interested in exploring the robustness of these parameters and their impact on model performance. Specifically, as illustrated in \Cref{fig:sen_lsm,fig:sen_alpha}, we observe that variations in both $\lambda$ and $\alpha$ do not result in significant fluctuations in the performance of \algo\ across three settings on CIFAR-100, indicating that the efficacy of \algo\ stems from its thoughtful design rather than an extensive reliance on parameter tuning. More analysis for hyperparameter sensitivity can be found in \Cref{app:param_sen}.

\vskip -0.15in
\begin{table}[h]
\centering
\caption{Results for PEFT strategies on \algo.}
\label{tab:app_peft}
\vskip 0.05in
\begin{small}
\begin{sc}
\resizebox{\columnwidth}{!}{
\begin{tabular}{@{}lccccc@{}}
\toprule
 &
  \multicolumn{3}{c}{CIFAR-100} &
  \multicolumn{2}{c}{Semi-Aves} \\ 
  \cmidrule(lr){2-4} \cmidrule(l){5-6}
Settings &
  N4 &
  N25 &
  N100 &
  $\mathcal{D}^{u}_{in}$ &
  $\mathcal{D}^{u}_{in} \cup \mathcal{D}^{u}_{out}$ \\ \midrule
\begin{tabular}[c]{@{}l@{}}VPT\\ Lora\\ Adapter\\ Adaptformer\end{tabular} &
  \begin{tabular}[c]{@{}c@{}}80.44\\ 79.71\\ 79.91\\ 81.02\end{tabular} &
  \begin{tabular}[c]{@{}c@{}}84.51\\ 85.56\\ 84.24\\ 85.44\end{tabular} &
  \begin{tabular}[c]{@{}c@{}}86.66\\ 87.29\\ 86.25\\ 87.50\end{tabular} &
  \begin{tabular}[c]{@{}c@{}}67.25\\ 68.30\\ 65.94\\ 68.39\end{tabular} &
  \begin{tabular}[c]{@{}c@{}}61.12\\ 61.13\\ 60.04\\ 60.73\end{tabular} \\ \bottomrule
\end{tabular}
}
\end{sc}
\end{small}
\vskip -0.05in
\end{table}

\noindent
\textbf{More PEFT Strategies.} While we use VPT for most experiments, it is noteworthy that various PEFT strategies are applicable. In \Cref{tab:app_peft}, Adaptformer \cite{chen2022adaptformer} exhibits an average performance surpassing VPT by up to 0.62\%. Another widely adopted PEFT strategy, Lora \cite{hu2021lora}, presents significant advantages in scenarios with a larger quantity of labeled data. Nevertheless, its performance is 0.73\% inferior to VPT in the N4 setting of CIFAR-100, which is characterized by a limited amount of labeled data. Adapter \cite{houlsby2019parameter} also performs reasonably well, yet they show an averaged 0.72\% decline in comparison to VPT. While it is acknowledged that Adaptformer yields superior performance in \Cref{tab:app_peft}, we choose VPT as the fine-tuning strategy in our primary results, considering its broader impact and potential applications \cite{han2024facing}.

\begin{table}[t]
\centering
\caption{The comparison for training time consumption (in seconds) on the CIFAR-100 dataset.}
\label{tab:time}
\vskip 0.05in
\begin{sc}
\resizebox{\columnwidth}{!}{%
\begin{tabular}{@{}lccc@{}}
\toprule
 &
  Per Step &
  \# Steps &
  Total \\ \midrule
\begin{tabular}[c]{@{}l@{}} FixMatch (Wide ResNet)\\ FixMatch (VPT) \\  \algo\ (VPT) \end{tabular} &
  \begin{tabular}[c]{@{}c@{}}0.053\\ 0.616\\ 0.642\end{tabular} &
  \begin{tabular}[c]{@{}c@{}}1024 $\times$ 1024\\ 30 $\times$ 500\\ 30 $\times$ 500\end{tabular} &
  \begin{tabular}[c]{@{}c@{}}55574.5\\ 9240.0\\ 9630.0\end{tabular} \\ \bottomrule
\end{tabular}%
}
\end{sc}
\vskip -0.15in
\end{table}

\noindent
\textbf{Time Consumption.} Our method stands out not just for its superior and robust performance but also for its training efficiency. \Cref{tab:time} shows that our proposed method \algo\ only requires about $1/6$ training time to converge compared with the consumption of training Wide ResNet from scratch. Additionally, FixMatch implemented in our framework, \textit{i.e.,} fine-tuning the foundation model via VPT, exhibits comparable time consumption with \algo\ which increases negligible training costs by introducing an auxiliary linear classifier to the original neural networks. The efficient framework we propose can be beneficial for promoting the deployment of SSL in real-world applications.

\begin{table}[t]
\centering
\caption{Ablation studies. We study the impact of core components of \algo\ on CIFAR-100 and Semi-Aves datasets.}
\label{tab:ablations}
\vskip 0.05in
\begin{sc}
\resizebox{1.0\linewidth}{!}{
\begin{tabular}{@{}lccccc@{}}
\toprule
 &
  \multicolumn{3}{c}{CIFAR-100} &
  \multicolumn{2}{c}{Semi-Aves} \\ \cmidrule(lr){2-4} \cmidrule(l){5-6}
Settings &
  N4 &
  N25 &
  N100 &
  $\mathcal{D}^{u}_{in}$ &
  $\mathcal{D}^{u}_{in} \cup \mathcal{D}^{u}_{out}$ \\ \midrule
\begin{tabular}[c]{@{}l@{}}\algo\\ w/o BMS \\ w/o Margin in $\ell^m_s$\\ w/o Dynamic $\alpha$\\ w/o DLS\\ w/o LS\\ w/o Detach\end{tabular} &
  \begin{tabular}[c]{@{}c@{}}80.44\\ 76.70\\ 80.52\\ 80.30\\ 80.29\\ 80.48\\ 79.85\end{tabular} &
  \begin{tabular}[c]{@{}c@{}}84.51\\ 83.71\\ 84.34\\ 84.34\\ 84.36\\ 84.40\\ 84.23\end{tabular} &
  \begin{tabular}[c]{@{}c@{}}86.66\\ 86.21\\ 86.60\\ 86.75\\ 86.36\\ 86.49\\ 86.78\end{tabular} &
  \begin{tabular}[c]{@{}c@{}}67.25\\ 65.83\\ 66.88\\ 67.10\\ 67.02\\ 67.22\\ 67.11\end{tabular} &
  \begin{tabular}[c]{@{}c@{}}61.12\\ 59.91\\ 58.94\\ 61.00\\ 58.58\\ 59.49\\ 61.05\end{tabular} \\ \bottomrule
\end{tabular}
}
\end{sc}
\vskip -0.15in
\end{table}

\begin{table*}[t]
\caption{Results for fine-tuning pre-trained ResNet CLIP on CIFAR-100.}
\label{tab:res}
\vskip 0.1in
\begin{center}
\begin{small}
\begin{sc}
\resizebox{0.85\textwidth}{!}{
\begin{tabular}{@{}lccccccccc@{}}
\toprule
Strategies & \multicolumn{3}{c}{BN Tuning} & \multicolumn{3}{c}{BitFit} & \multicolumn{3}{c}{SSF} \\ \cmidrule(l){2-10} 
Settings   & N4       & N25      & N100    & N4        & N25      & N100     & N4     & N25    & N100  \\ \midrule
\begin{tabular}[c]{@{}l@{}}FixMatch\\ FlexMatch\\ FreeMatch\\ SoftMatch\\ DebiasPL\end{tabular} &
  \begin{tabular}[c]{@{}c@{}}37.49\\ 38.22\\ 30.51\\ 28.70\\ \underline{42.72}\end{tabular} &
  \begin{tabular}[c]{@{}c@{}}\underline{62.98}\\ 51.96\\ 61.48\\ 61.32\\ 60.62\end{tabular} &
  \begin{tabular}[c]{@{}c@{}}\underline{71.48}\\ 69.52\\ 70.11\\ 68.80\\ 69.76\end{tabular} &
  \begin{tabular}[c]{@{}c@{}}52.67\\ \textbf{55.69}\\ 53.28\\ 53.34\\ 52.90\end{tabular} &
  \begin{tabular}[c]{@{}c@{}}66.15\\ 66.33\\ 66.52\\ 66.46\\ \underline{66.62}\end{tabular} &
  \begin{tabular}[c]{@{}c@{}}70.19\\ 69.58\\ 69.07\\ 69.38\\ \underline{70.34}\end{tabular} &
  \begin{tabular}[c]{@{}c@{}}36.49\\ 36.42\\ 36.29\\ 36.11\\ \underline{38.83}\end{tabular} &
  \begin{tabular}[c]{@{}c@{}}\underline{53.47}\\ 51.89\\ 52.01\\ 52.09\\ 52.54\end{tabular} &
  \begin{tabular}[c]{@{}c@{}}\textbf{60.44}\\ 57.49\\ 57.35\\ 57.51\\ 58.44\end{tabular} \\ \midrule
\algo\    & \textbf{48.32}    & \textbf{65.87}    & \textbf{72.38}   & \underline{55.21}     & \textbf{67.06}    & \textbf{70.60}    & \textbf{39.40}  & \textbf{54.14}  & \underline{59.51} \\ \bottomrule
\end{tabular}%
}
\end{sc}
\end{small}
\end{center}
\end{table*}

\subsection{Ablation Analysis}
\label{sec:ablation}

To better understand \algo, we tease apart the factors that contribute significantly to its success in \Cref{tab:ablations}.

\noindent
\textbf{Impact of balanced margin softmax.} The balanced margin softmax loss (denoted by ``BMS'' in \Cref{tab:ablations}) serves as a crucial component for alleviating {aggregated biases} in our method. Upon removing the dynamic adjustments to the margins, we notice a significant performance decline across all settings, and particularly noteworthy is the substantial 3.74\% decrease in performance observed in the ``N4'' setting on CIFAR-100, where the labeled data is quite limited. This shows that BMS can effectively erase model biases.

\noindent
\textbf{Impact of balanced margin softmax in supervised loss.} In our method, we also incorporate dynamically computed margins based on unlabeled data into the labeled loss $\mathcal{L}^m_s$, thereby enhancing the balance and stability of the model. The exclusion of the margins introduced in $\mathcal{L}^m_s$ resulted in an average performance drop of 1.28\% on Semi-Aves, emphasizing the necessity of incorporating the margins into the labeled loss.


\noindent
\textbf{Impact of dynamic scaling parameter $\alpha$.} 
The dynamic scaling parameter $\alpha_t$ in \Cref{eq:alpha} plays a crucial role in adjusting the magnitude of class margins. To examine its impact, we set $\alpha_t$ to a constant value in this experiment. From \Cref{tab:ablations}, we observe a decrease in accuracy across most settings. This suggests that the dynamic adjustment of $\alpha$ is essential for improving the model's ability to handle imbalanced pseudo-label distributions.

\noindent
\textbf{Impact of decoupled label smoothing.} In this experiment, we remove the auxiliary classifier and reweight unlabeled data using confidence scores generated by the main branch. We observe a consistent performance drop across all settings, averaging 0.67\%. Especially in the presence of OOD data on Semi-Aves, the performance decreases by 2.54\%. This highlights that discriminative model probabilities play an important role in sample reweighting.

\noindent
\textbf{Impact of label smoothing in the auxiliary classifier.} We employ label smoothing for learning the auxiliary classifier to address {cognitive deviation}. Removing label smoothing, which means training the auxiliary classifier using hard pseudo-labels directly, leads to performance deterioration in most settings. Particularly, we observe a 1.63\% decrease in accuracy on Semi-Aves when the unlabeled data contains OOD samples.

\noindent
\textbf{Impact of the auxiliary classifier on representations.} 
In \algo, the auxiliary classifier is detached to prevent hurting the representation learning. We investigate the impact of the auxiliary branch on representation learning by allowing gradient backpropagation to the former layers of the model. The results are denoted by ``w/o Detach'' in \Cref{tab:ablations}. We observe a substantial performance drop when the number of labeled samples is extremely scarce, \textit{e.g.,} a 0.59\% decrease in the ``N2'' setting on CIFAR-100. This shows that mitigating overconfidence may hurt representations.

\begin{figure}[t]
  \centering
  \begin{subfigure}[b]{0.238\textwidth}
  \centering
  \includegraphics[height=3.15cm,width=\linewidth]{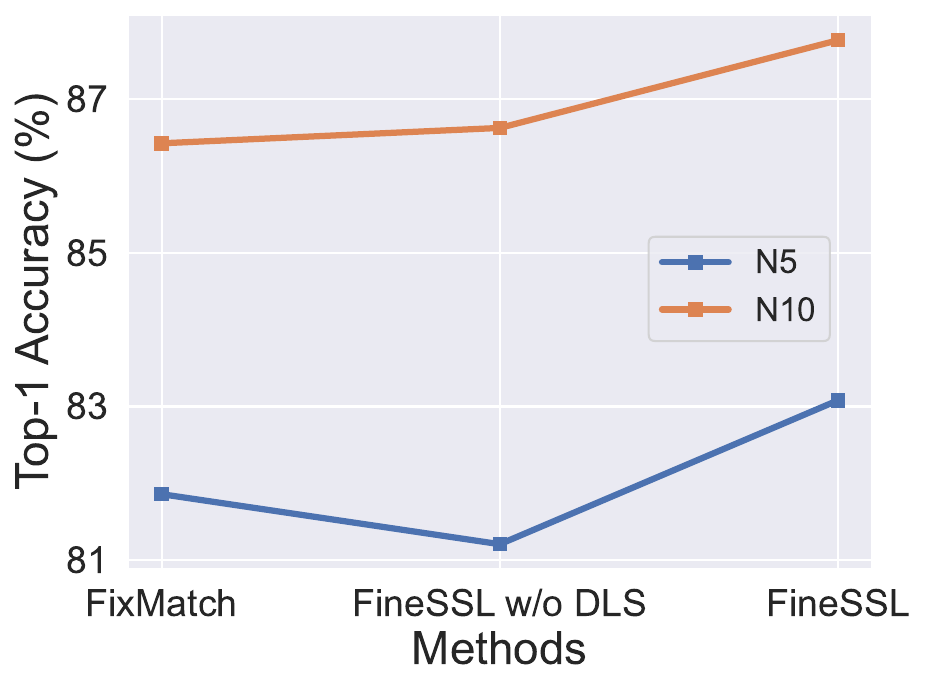}
  \caption{}
  \label{fig:ood_acc}
  \end{subfigure}
 \begin{subfigure}[b]{0.238\textwidth}
  \centering
  \includegraphics[height=3.1cm,width=\linewidth]{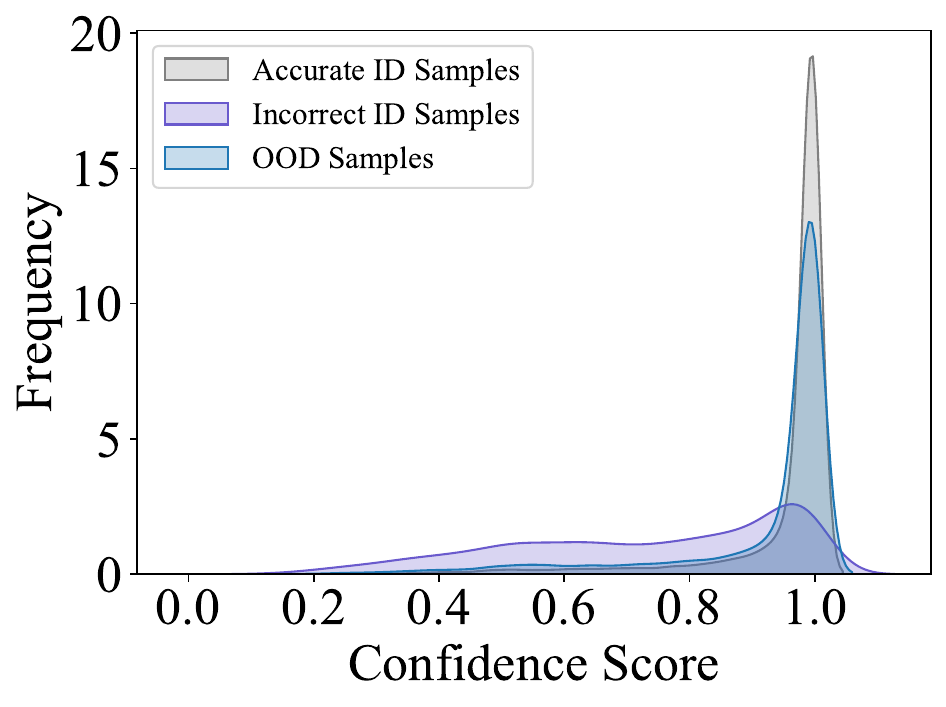}
  \caption{}
  \label{fig:ood1}
 \end{subfigure}

 \medskip

\begin{subfigure}[b]{0.238\textwidth}
  \centering
  \includegraphics[height=3.15cm,width=\linewidth]{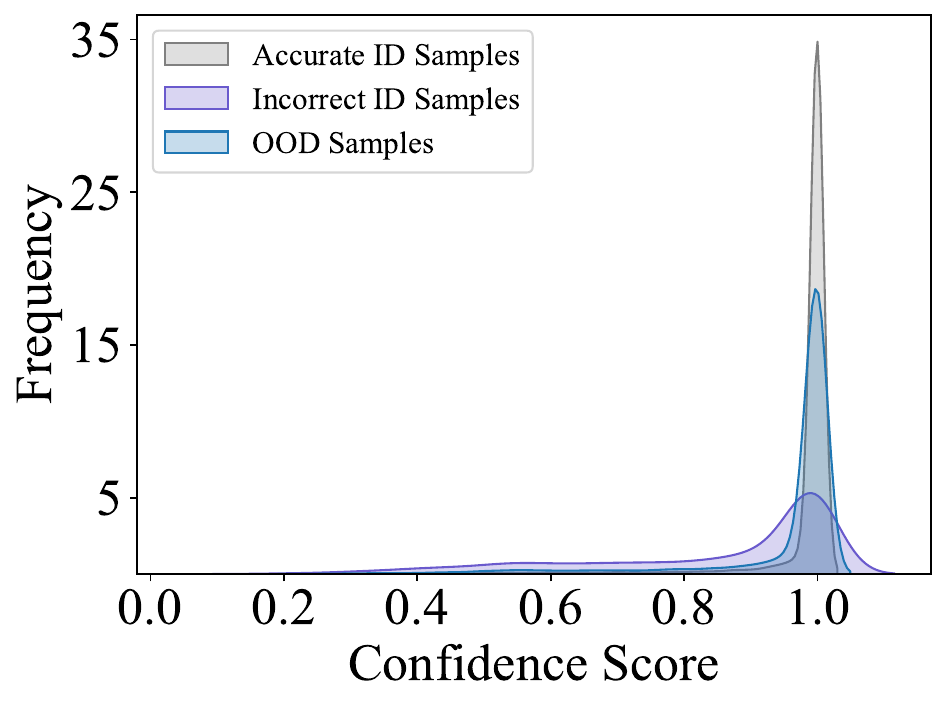}
  \caption{}
  \label{fig:ood2}
  \end{subfigure}
 \begin{subfigure}[b]{0.238\textwidth}
  \centering
  \includegraphics[height=3.1cm,width=\linewidth]{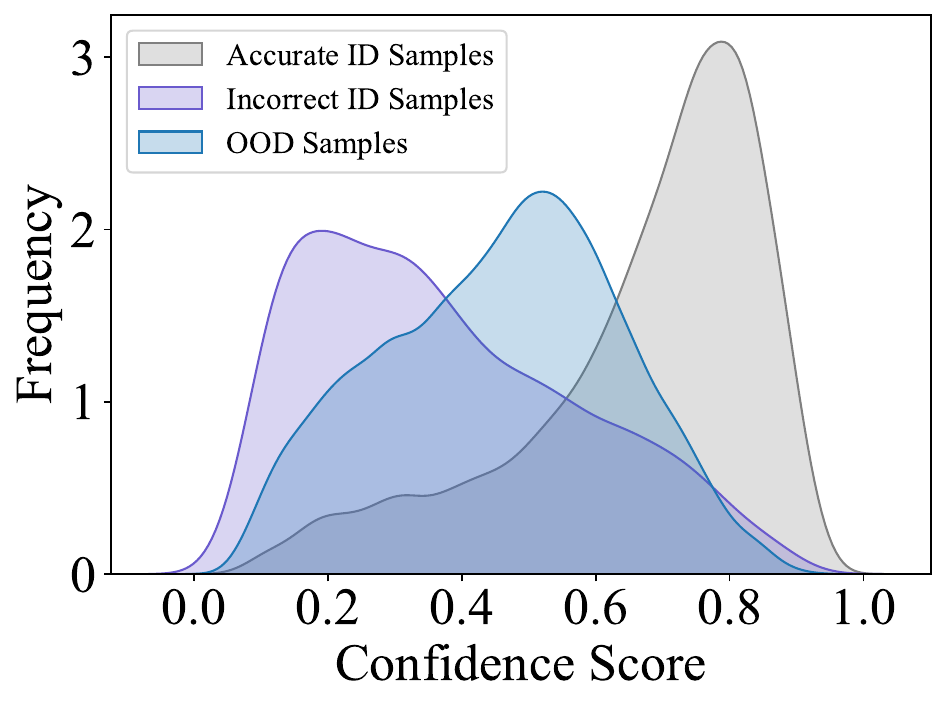}
  \caption{}
  \label{fig:ood3}
 \end{subfigure}
 
  \vskip -0.1in
  \caption{(\labelcref{fig:ood_acc}): The accuracy for \textit{N5} and \textit{N10} OpenSSL setting on CIFAR-100 for different methods. (\labelcref{fig:ood1,fig:ood2,fig:ood3}): The distribution of confidence score for ID and OOD samples of FixMatch, \algo\ w/o DLS, and \algo\ for \textit{N5} OpenSSL setting.}
  \label{fig:ood}
\end{figure}

\subsection{Tasks with OOD Samples}
\label{sec:ood_samples}

In \Cref{sec:ablation}, we observe that DLS exhibits significant performance improvement on Semi-Aves with OOD samples. Notably, the presence of OOD samples is prevalent in SSL \cite{cao2022openworld,guo2022nach}. Motivated by this finding, we conduct further experiments to investigate the role of DLS in handling OOD samples. Specifically, following OpenSSL settings in \cite{cao2022openworld}, we divide the classes on CIFAR-100 into 50\% seen classes (ID samples) and 50\% novel classes (OOD samples). We select 5 and 10 labeled samples per seen class (i.e., N5 and N10 settings), with the remaining samples unlabeled. We show the accuracy for seen classes of the test set in \Cref{fig:ood_acc}. The results clearly demonstrate that removing the DLS leads to an average performance decline of 1.87\%, indicating the effectiveness of DLS in mitigating the impact of OOD samples.

\noindent
\textbf{Why DLS can benefit tasks with OOD samples?} We show the confidence score distribution of accurate, incorrect, and OOD samples in \Cref{fig:ood1,fig:ood2,fig:ood3}. For FixMatch and \algo\ w/o DLS, we observe significant overlap in the distribution regions of the three types of samples, suggesting that the model faces challenges in distinguishing them using confidence scores. However for \algo, there exists much more clearer confidence boundary, which enables the model to assign more reliable weights to each sample, leading to performance improvements for ID and OOD samples.

\subsection{Results for pre-trained ResNet model}
\label{sec:resnet}
Although this paper primarily focuses on improving the CLIP model based on ViT structure, we still conduct experiments using the pre-trained ResNet CLIP model. As VPT and other PEFT strategies are not applicable to ResNet, we employ BN Tuning (fine-tuning batchnorm parameters), BitFit \cite{zaken2021bitfit} (fine-tuning solely the bias terms), and SSF \cite{lian2022scaling} for fine-tuning. The results in \Cref{tab:res} consistently demonstrate the superiority of \algo\ over other comparative methods. Notably, \algo\ achieves an average improvement of 2.19\% compared to DebiasPL, highlighting its significant performance advantages in fine-tuning the ResNet structure model. 

\vspace{-0.05in}
\section{Conclusion}
This paper proposes a novel SSL approach \algo\ to improve the generalization performance by efficiently fine-tuning pre-trained foundation models. We identify the inherent aggregated biases and cognitive deviation issues in the foundation model, which hinder the generation and selection of reliable pseudo-labels. We introduce the balanced margin softmax to erase the model biases across different classes. Further, to take full advantage of unlabeled data, we devise a decoupled label smoothing to regularize model confidence and reweight unlabeled samples. Extensive experimental results on five public datasets demonstrate that \algo\ achieves superior performance over the previous state of the arts and decreases the training cost by six times. Moreover, The framework we implemented can smoothly incorporate many fine-tuning and SSL methods.

\section*{Impact Statement}
This paper presents work whose goal is to advance the field of machine learning. Specifically, we propose a new semi-supervised learning algorithm to improve the performance on low-label regimes by fine-tuning pre-trained foundation models. In classification tasks, since pre-trained models can be biased toward certain groups of classes, this paper makes a substantial attempt to erase such inherent prediction biases. There are many potential societal consequences of our work, none of which we feel must be specifically highlighted here.


\bibliography{example_paper}
\bibliographystyle{icml2024}

\newpage
\appendix
\onecolumn

\section{Implementation Details}
\label{app:implementation}

\subsection{Implementation Details}
\label{app:exp_details}
To ensure the reproducibility of our method's performance, we present experimental details below, and default configs for the experiments are shown in \Cref{tab:exp_details}.
For most settings, a batch size of 32 is utilized. However, due to the limited availability of labeled data in the CIFAR-10 ``N1'' and ``N2'' settings, batch sizes of 8 and 16 are chosen, respectively. The relative batch size ratio $\mu$ between unlabeled and labeled data is set to 2 for the ImageNet dataset, and it is consistently set to 1 for the remaining datasets. Note that training is performed for 30 epochs with each epoch consisting of 500 steps, which is significantly less than the 1024 $\times$ 1024 steps employed by previous training from scratch methods \cite{sohn2020fixmatch,wang2022freematch}. For certain datasets, such as ImageNet, additional training epochs and steps are expected to yield further performance improvements, which we intend to explore in the future. For competing methods implemented in our general fine-tuning framework for SSL, we find that the performance achieved with a higher threshold, specifically 0.95, is unsatisfactory. Hence, in our experiments, we adopt a smaller threshold of 0.7 for them, which can achieve superior results.

\vskip -0.1in
\begin{table}[h]
\caption{The default parameter configs employed in the experiments.}
\label{tab:exp_details}
\vskip 0.1in
\begin{center}
\begin{tabular}{@{}cc@{}}
\toprule
 Configuration &
  Default Value \\ \midrule
\begin{tabular}[c]{@{}c@{}}Optimizer\\ Learning rate\\ Scheduler for lr\\ Weight decay\\ Momentum factor\\ Batch Size\\ Model\\ Epochs\\ Steps\\ $\mu$\\ PEFT strategy\\ VPT length\\ $\lambda$\\ $\alpha$\\ $\gamma$\end{tabular} &
  \begin{tabular}[c]{@{}c@{}}SGD\\ 0.03\\ Cosine decay\\ $5 \times 10^{-4}$\\ 0.9\\ 32\\ CLIP-ViT\\ 30\\ 500\\ 1\\ VPT deep\\ 50\\ 0.5\\ 8.0\\ 3.0\end{tabular} \\ \bottomrule
\end{tabular}
\end{center}
\end{table}


\section{Additional Experiments}

\subsection{Long-Tailed Semi-Supervised Learning}

Most of the experiments outlined above assume a uniform distribution of labeled and unlabeled data, whereas data in the real world often exhibits a long-tailed distribution. Therefore, we integrate some long-tailed semi-supervised learning (LTSSL) settings to further validate the effectiveness of \algo. It is noteworthy that, given our discovery of the existence of {aggregated biases}, combining long-tailed data distributions makes these settings particularly challenging.

In \Cref{tab:long_tailed}, $N1$ and $M_1$ mean the number of samples for the first class in labeled and unlabeled data, and typically, we have $N_1 \ge N_2 \ge \dots \ge N_C$ and $M_1 \ge M_2 \ge \dots \ge M_C$. We use $\rho$ to represent the imbalance ratio for labeled and unlabeled data, where $\rho = \frac{N_1}{N_C}$ and $N_k = N_1 \times \rho^{-\frac{k - 1}{C - 1}}$ for class $k \in [1, C]$. The results indicate that \algo\ exhibits a significant advantage over all comparative methods, averaging a 2.95\% performance improvement over SoftMatch, which suggests the superior robustness of our method against data imbalance. It is noteworthy that DebiasPL also demonstrates favorable performance in LTSSL settings, which is attributed to its incorporation of techniques specifically addressing long-tailed challenges \cite{menon2020long}. Nevertheless, \algo\ still surpasses DebiasPL across the majority of settings.

\begin{table}[ht]
\caption{Comparison for long-tailed semi-supervised learning settings on CIFAR-100 dataset.}
\label{tab:long_tailed}
\vskip 0.1in
\begin{center}
\begin{small}
\begin{sc}
\begin{tabular}{@{}lcccc@{}}
\toprule
Settings        & \multicolumn{2}{c}{\begin{tabular}[c]{@{}c@{}}$N_1=50$\\ $M_1=400$\end{tabular}} & \multicolumn{2}{c}{\begin{tabular}[c]{@{}c@{}}$N_1=150$\\ $M_1=300$\end{tabular}} \\
\cmidrule(lr){2-3} \cmidrule(lr){2-3}\cmidrule(l){4-5}
$\rho$ & 10                                 & 20                                & 10                                 & 20                                 \\ \midrule
\begin{tabular}[c]{@{}l@{}}FixMatch\\ FlexMatch\\ FreeMatch\\ SoftMatch\\ DebiasPL\end{tabular} &
  \begin{tabular}[c]{@{}c@{}}78.94\\ 78.81\\ 79.05\\ 78.42\\ \underline{81.60}\end{tabular} &
  \begin{tabular}[c]{@{}c@{}}72.04\\ 73.36\\ 72.82\\ 72.39\\ \textbf{78.23}\end{tabular} &
  \begin{tabular}[c]{@{}c@{}}81.63\\ 81.75\\ 81.49\\ 81.52\\ \underline{83.12}\end{tabular} &
  \begin{tabular}[c]{@{}c@{}}78.21\\ 78.53\\ 78.42\\ 78.90\\ \underline{80.89}\end{tabular} \\ \midrule
Ours            & \textbf{81.63}                              & \underline{77.75}                             & \textbf{83.44}                              & \textbf{81.35}                              \\ \bottomrule
\end{tabular}
\end{sc}
\end{small}
\end{center}
\end{table}

\subsection{Can FFT Behave More Effective?}

In \Cref{fig:res_vit_bar}, we observe that FFT achieves inferior performance in semi-supervised learning settings, particularly when labeled data is extremely scarce. However, we find that a simple modification can enhance the performance of FFT. Specifically, \cite{kumar2022fine} argues that a two-step strategy of linear probing and then full fine-tuning (LP-FT), can achieve great improvement in ID and OOD accuracy. Interestingly, LP-FT can also achieve excellent performance in SSL settings. From \Cref{tab:ft_lp}, FixMatch with LP-FT can achieve an averaged 48.99\% enhancement in performance compared to the FFT of FixMatch.

In addition, following PEL \cite{shi2023parameter}, we find that semantic-aware initialization for the linear classifier can also help FFT achieve better performance. Specifically, we employ textual features associated with class labels to initialize the classifier weights. This straightforward strategy facilitates the classifier to attain a more optimal initial state without incurring additional computational overhead, which accelerates model convergence and maintains a more stable training process. In \Cref{tab:ft_lp}, the performance of FFT$^{\dagger}$ closely approximates that of VPT, exhibiting an average superiority of 47.30\% over FFT.
It is noteworthy that we search the learning rate from \{0.0005, 0.0003, 0.0001, 0.00005, 0.00003, 0.00001\} and choose the best performance for all experiments related to FFT. In addition, it is obvious that \algo\ outperforms FixMatch across all fine-tuning strategies listed in \Cref{tab:ft_lp}, averaging 4.59\%, indicating the broad applicability and effectiveness of our method.

Overall, LP-FT and FFT$^{\dagger}$ both yield a well-suited weight initialization for the classifier, which aids in stabilizing the model during training. The guidance provided by a well-initialized classifier prevents the model from collapsing, thereby avoiding low performance. We suggest that the question of why initialization assists in achieving superior performance in FFT deserves more attention and research in future work to reveal more profound reasons.

\vskip -0.1in
\begin{table}[ht]
\caption{More results for LP and FFT on CIFAR-100 dataset. FFT$^{\dagger}$ denotes using semantic-aware initialization.}
\label{tab:ft_lp}
\vskip 0.1in
\begin{center}
\begin{small}
\begin{sc}
\begin{tabular}{@{}llll@{}}
\toprule
Settings &
  \multicolumn{1}{c}{N4} &
  \multicolumn{1}{c}{N25} &
  \multicolumn{1}{c}{N100} \\ \midrule
\begin{tabular}[c]{@{}l@{}}FixMatch w/ LP\\ \algo\ w/ LP\end{tabular} &
  \multicolumn{1}{c}{\begin{tabular}[c]{@{}c@{}}63.64\\ 68.22\end{tabular}} &
  \multicolumn{1}{c}{\begin{tabular}[c]{@{}c@{}}73.66\\ 75.57\end{tabular}} &
  \multicolumn{1}{c}{\begin{tabular}[c]{@{}c@{}}78.01\\ 78.59\end{tabular}} \\ \midrule
\begin{tabular}[c]{@{}l@{}}FixMatch w/ FFT \\ \algo\ w/ FFT\end{tabular} &
  \begin{tabular}[c]{@{}l@{}}6.56\\ 8.07\end{tabular} &
  \begin{tabular}[c]{@{}l@{}}32.93\\ 41.64\end{tabular} &
  \begin{tabular}[c]{@{}l@{}}49.76\\ 59.88\end{tabular} \\ \midrule
\begin{tabular}[c]{@{}l@{}}FixMatch w/ LP-FT\\ \algo\ w/ LP-FT\end{tabular} &
  \begin{tabular}[c]{@{}l@{}}70.80\\ 77.45\end{tabular} &
  \begin{tabular}[c]{@{}l@{}}80.25\\ 83.18\end{tabular} &
  \begin{tabular}[c]{@{}l@{}}85.16\\ 86.04\end{tabular} \\ \midrule
\begin{tabular}[c]{@{}l@{}}FixMatch w/ FFT$^{\dagger}$\\ \algo\ w/ FFT$^{\dagger}$\end{tabular} &
  \begin{tabular}[c]{@{}l@{}}67.77\\ 78.03\end{tabular} &
  \begin{tabular}[c]{@{}l@{}}79.82\\ 84.04\end{tabular} &
  \begin{tabular}[c]{@{}l@{}}83.56\\ 86.33\end{tabular} \\ \bottomrule
\end{tabular}%
\end{sc}
\end{small}
\end{center}
\end{table}

\subsection{ImageNet-21k Pre-Trained Model}

Our experiments predominantly focus on the pre-trained ViT by initializing weights using the image encoder of CLIP. Here we delve into the performance of the model pre-trained with weights derived from ImageNet-21k \cite{deng2009imagenet}. From \Cref{tab:ink}, we can see that the fine-tuning of the model pre-trained on ImageNet-21k yields superior performance compared to the CLIP pre-trained model. We assert that this phenomenon can be attributed to the substantial correlation and similarity in image features between the ImageNet-21k and CIFAR-100 as well as Semi-Aves. This further clarifies why we predominantly employ the CLIP as the pre-trained model in most experiments, given its broader semantic feature space and substantially reduced feature similarity between pre-training and downstream training data. In addition, the ViT from Clip is trained using contrastive loss, which might bring a gap for direct downstream fine-tuning for image classification. Examining the specific numerical results in \Cref{tab:ink}, \algo\ continues to show its robust performance superiority, surpassing DebiasPL by an average of 0.36\%.

\vskip -0.1in
\begin{table}[ht]
\caption{Results for fine-tuning the foundation model pre-trained on ImageNet-21k dataset.}
\label{tab:ink}
\vskip 0.1in
\begin{center}
\begin{small}
\begin{sc}
\begin{tabular}{@{}lccccc@{}}
\toprule
  & \multicolumn{3}{c}{CIFAR-100} & \multicolumn{2}{c}{Semi-Aves} \\ 
\cmidrule(lr){2-4} \cmidrule(l){5-6}
Settings & N4       & N25      & N100    & $\mathcal{D}^{u}_{in}$           & $\mathcal{D}^{u}_{in} \cup \mathcal{D}^{u}_{out}$          \\ \midrule
\begin{tabular}[c]{@{}l@{}}FixMatch\\ FlexMatch\\ FreeMatch\\ SoftMatch\\ DebiasPL\end{tabular} &
  \begin{tabular}[c]{@{}c@{}}87.05\\ 88.99\\ 82.82\\ 82.93\\ \underline{89.96}\end{tabular} &
  \begin{tabular}[c]{@{}c@{}}90.69\\ 90.95\\ 90.76\\ 90.99\\ \underline{91.00}\end{tabular} &
  \begin{tabular}[c]{@{}c@{}}91.43\\ \textbf{91.88}\\ 91.70\\ 91.59\\ 91.57\end{tabular} &
  \begin{tabular}[c]{@{}c@{}}67.89\\ 66.63\\ 66.86\\ 66.55\\ \underline{69.40}\end{tabular} &
  \begin{tabular}[c]{@{}c@{}}63.63\\ 61.25\\ 61.36\\ 61.34\\ \underline{63.93}\end{tabular} \\ \midrule
Ours     & \textbf{90.30}    & \textbf{91.24}    & \underline{91.87}   & \textbf{70.01}         & \textbf{64.25}         \\ \bottomrule
\end{tabular}
\end{sc}
\end{small}
\end{center}
\end{table}

\subsection{Performance comparisons for ViT training from scratch}

Previous work \cite{weng2022semi} has found that training ViTs from scratch fails to yield satisfactory performance on SSL tasks, particularly with limited labeled data. Furthermore, the significantly longer training time required for training ViTs from scratch is often impractical in real-world scenarios. 

To validate this, we conduct experiments comparing the performance of training a ViT-small from scratch using FixMatch for 500 epochs, which is significantly longer than the 30 epochs used for VPT. Actually, ViT-base tends to show inferior performance due to more parameters \cite{wang2022usb}, which provides reason for our adoption of ViT-small. The performance of training from scratch is observed to be considerably poorer, with a 54.77\% decrease compared to fine-tuning by VPT. 

Additionally, we include a comparison between our proposed method, \algo\, and Semiformer \cite{weng2022semi}, a method specifically designed for SSL tasks with ViTs. We focus on comparing results for the ImageNet dataset with 10\% labeled data. The results demonstrate a 3.71\% performance advantage of \algo\ over Semiformer in this setting, indicating significant advantages over training from scratch.

\subsection{Additional comparisons with previous research}
We incorporate thorough comparison with the strategies Few-PseudoLabels (FPL), Iterative Refinement of FPL (IFPL), and Grow and Refine Iteratively Pseudolabels (GRIP) as proposed in \cite{menghini2023enhancing}. To ensure fair comparisons, we compare these strategies in both single-modality settings using VPT and multiple-modality settings using CoOp \cite{zhou2022learning} and VPT.

Our results show that \algo\ exhibits significant performance advantages, particularly in scenarios with limited labeled data, such as N4 and N25. It is worth noting that the use of a large volume of unlabeled data by FPL and IFPL is constrained due to the selection of a fixed number of unlabeled samples for each class. Moreover, the selected unlabeled data primarily consists of easy samples that are relatively simple to classify, thereby contributing only marginally to performance improvement. Furthermore, aggregated biases can negatively impact GRIP, resulting in substantial imbalance in pseudo-labels.

\begin{table}[ht]
\caption{Results for FPL, IFPL, and GRIP in three settings (N4 / N25 / N100) on CIFAR-100 dataset.}
\label{tab:enhance_clip}
\vskip 0.1in
\begin{center}
\begin{small}
\begin{sc}
\begin{tabular}{@{}lccc@{}}
\toprule
Dataset &
  \multicolumn{3}{c}{CIFAR-100} \\ \midrule
Settings &
  N4 &
  N25 &
  N100 \\ \midrule
\begin{tabular}[c]{@{}l@{}}FPL w/ VPT\\ IFPL w/ VPT\\ GRIP w/ VPT\\ FPL w/ VPT + CoOp\\ IFPL w/ VPT + CoOp\\ GRIP w/ VPT + CoOp\\ \algo\ \end{tabular} &
  \begin{tabular}[c]{@{}c@{}}74.80\\ 74.71\\ 70.24\\ 74.94\\ 74.72\\ 68.65\\ \textbf{80.44}\end{tabular} &
  \begin{tabular}[c]{@{}c@{}}80.23\\ 80.93\\ 80.83\\ 80.72\\ 81.10\\ 79.94\\ \textbf{84.51}\end{tabular} &
  \begin{tabular}[c]{@{}c@{}}84.24\\ 84.28\\ 84.32\\ 84.49\\ 84.69\\ 84.17\\ \textbf{86.66}\end{tabular} \\ \bottomrule
\end{tabular}%
\end{sc}
\end{small}
\end{center}
\end{table}

\section{Further Analysis on \algo}
\label{app:further_ana}

\subsection{Sensitivity Studies for Hyperparameters}
\label{app:param_sen}

\begin{figure}[ht]
\begin{center}
   \begin{subfigure}[b]{0.31\textwidth}
   \centering
   \includegraphics[width=\linewidth]{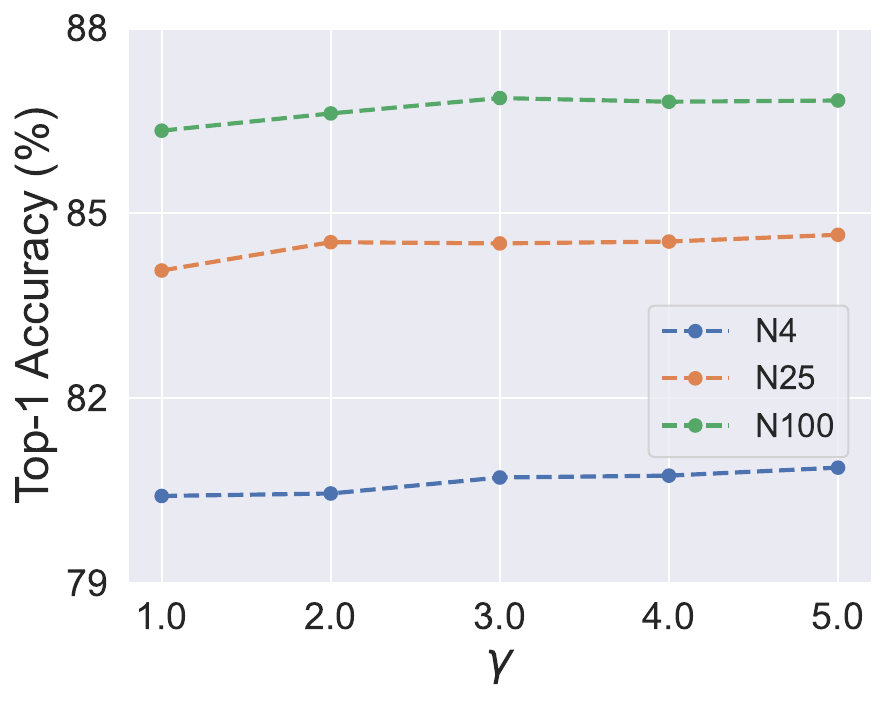}
   \caption{}
   \label{fig:sen_gamma}
   \end{subfigure}
   \begin{subfigure}[b]{0.31\textwidth}
   \centering
   \includegraphics[width=\linewidth]{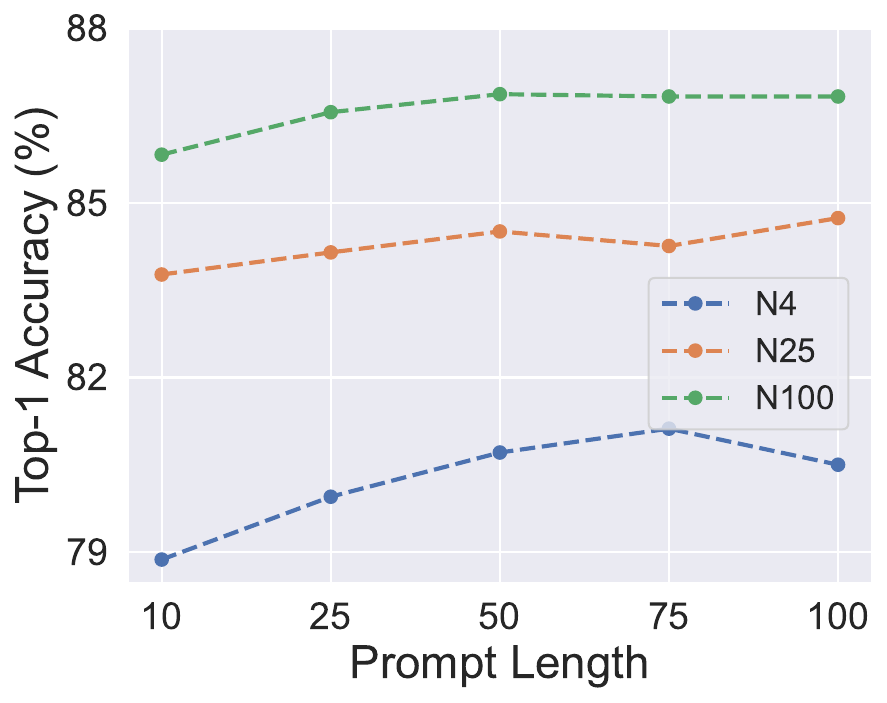}
   \caption{}
   \label{fig:sen_vptlen}
   \end{subfigure}
   \begin{subfigure}[b]{0.31\textwidth}
   \centering
   \includegraphics[width=\linewidth]{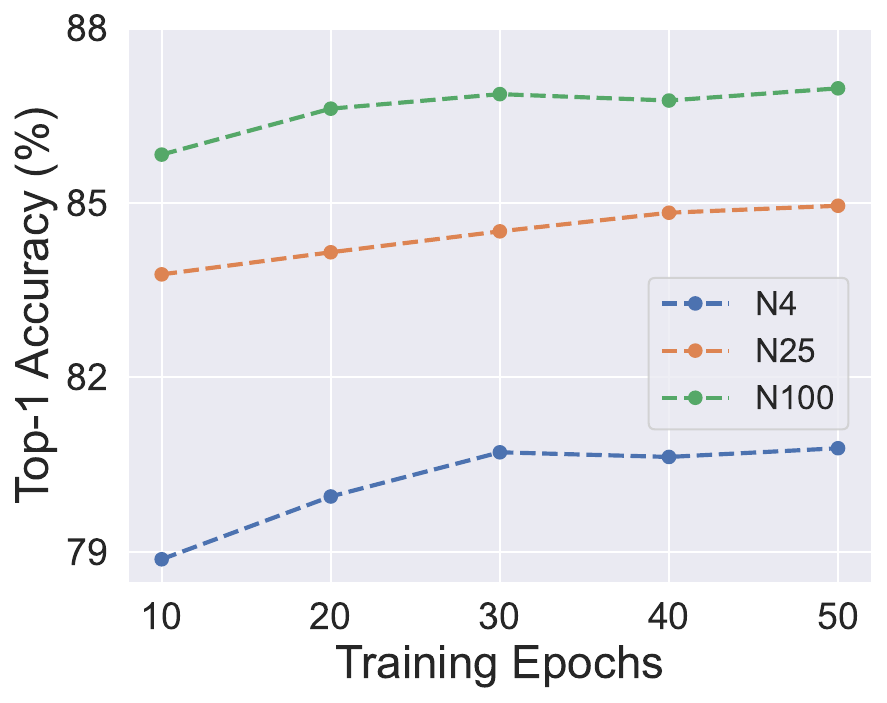}
   \caption{}
   \label{fig:sen_epochs}
   \end{subfigure}
  \vskip -0.1in
  \caption{(\labelcref{fig:sen_gamma,fig:sen_vptlen,fig:sen_epochs}): The sensitivity for $\gamma$, prompt length and training epochs under various settings on CIFAR-100.}
\label{fig:app_sen}
\end{center}
\vskip -0.1in
\end{figure}

First, we investigate the impact of $\gamma$, which is a hyperparameter employed to scale the sample weights when calculating the consistency regularizer in \Cref{eq:ulab_weight}. From \Cref{fig:sen_gamma}, we can see that changing the value of $\gamma$ does not lead to substantial fluctuations in the generalization performance, indicating the robustness of our method. 

Next, we study the influence of prompt length for VPT in \Cref{fig:sen_vptlen}. We observe a marginal performance enhancement as the VPT length increases, which means the performance can be further improved by adding more learnable model parameters. In our implementation, we set the VPT length to 50, which balances the effectiveness and efficiency.

Finally, \Cref{fig:sen_epochs} depicts the performance curve as a function of the number of training epochs. Generally, the test accuracy continues to rise as the training progresses. However, the performance stabilizes after training for 30 epochs. Therefore, we choose to fine-tune the model for 30 epochs in our experiments, which is adequate to obtain sufficiently well performance.

\subsection{Alleviation of Aggregated Biases and Cognitive Deviation}

\begin{figure}[ht]
  \centering
   \begin{subfigure}[b]{0.3\textwidth}
   \centering
   \includegraphics[width=\linewidth]{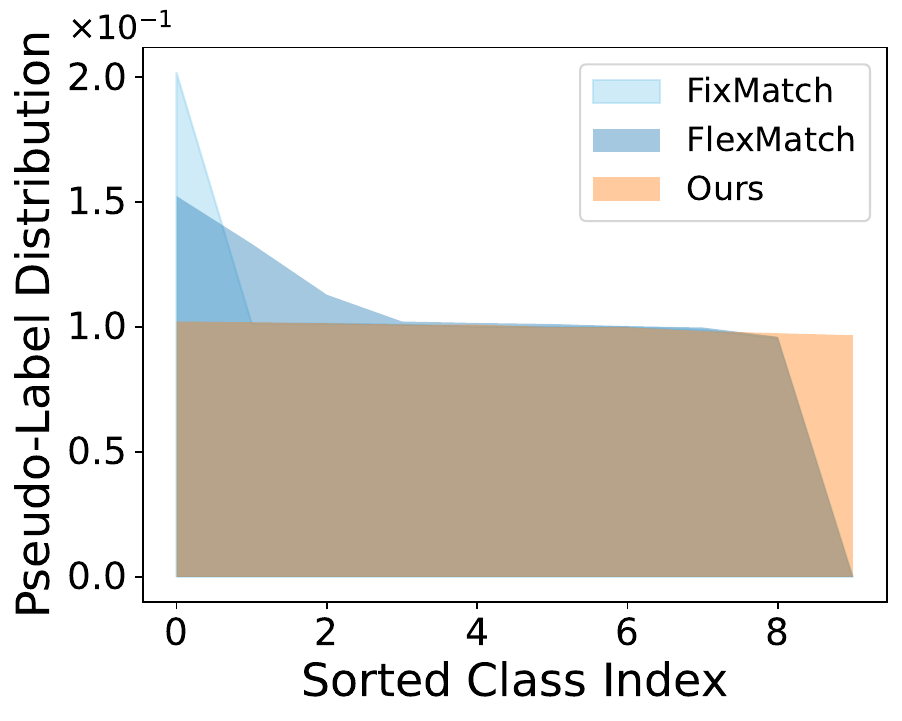}
   \caption{ CIFAR-10}
   \label{fig:area_ulab_cifar10}
   \end{subfigure}
   \hspace{2em}
   \begin{subfigure}[b]{0.29\textwidth}
   \centering
   \includegraphics[width=\linewidth]{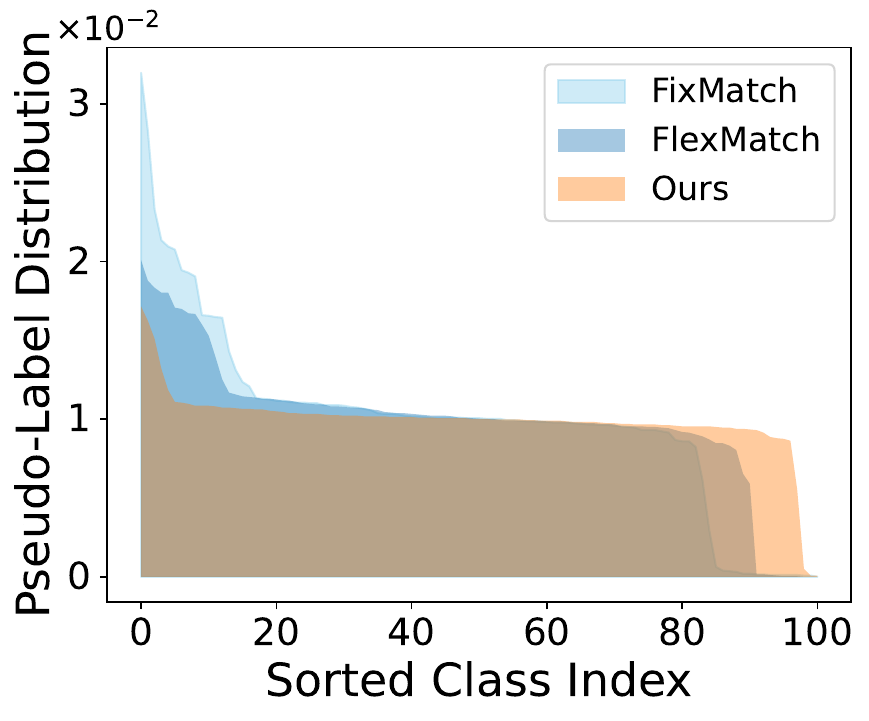}
   \caption{FOOD-101}
   \label{fig:area_ulab_food101}
   \end{subfigure}
   \vskip -0.1in
   \caption{Distribution of pseudo-labels for FixMatch, FlexMatch, and \algo\ in the ``N2'' setting.}
\end{figure}

\begin{figure}[ht]
\centering
   \begin{subfigure}[b]{0.3\textwidth}
   \centering
   \includegraphics[width=\linewidth]{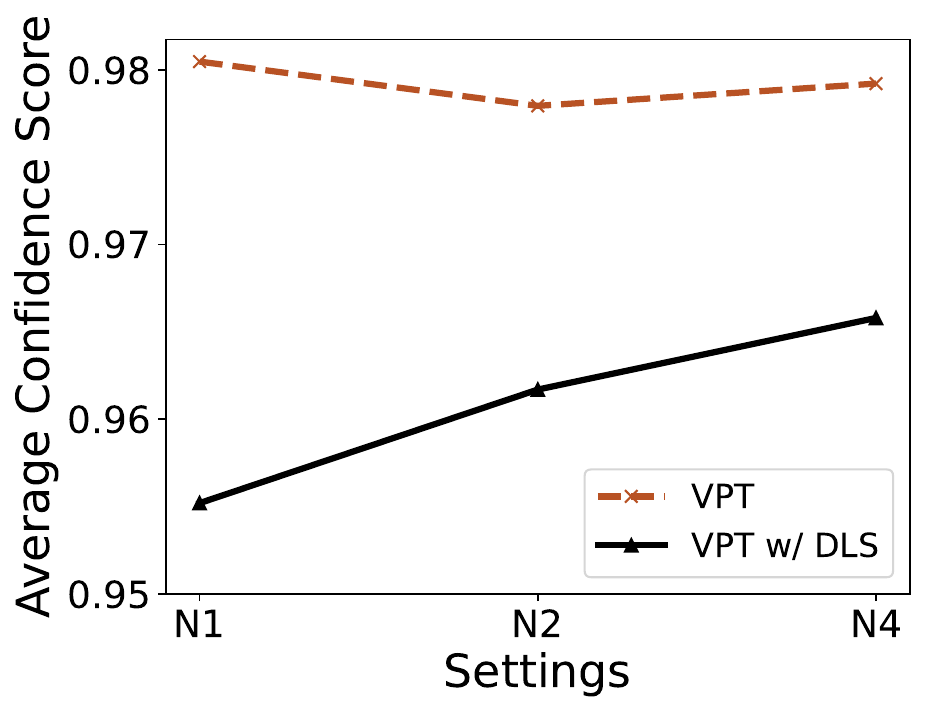}
   \caption{CIFAR-10}
   \label{fig:line_conf_cifar10}
   \end{subfigure}
   \hspace{2em}
   \begin{subfigure}[b]{0.3\textwidth}
   \centering
   \includegraphics[width=\linewidth]{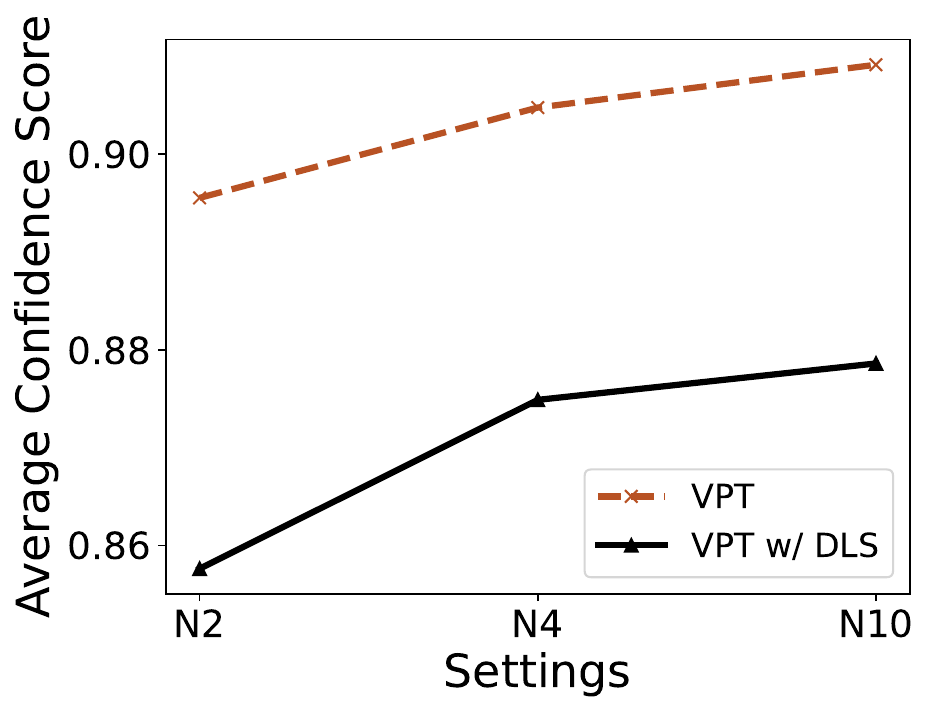}
   \caption{FOOD-101}
   \label{fig:line_conf_food101}
   \end{subfigure}
   \vskip -0.1in
  \caption{Average confidence in tasks of different numbers of labeled samples for models with and without DLS.}
\label{fig:app_ana}
\vskip -0.1in
\end{figure}

We present additional figures to depict the alleviating effects on {aggregated biases} and {cognitive deviation} in \Cref{fig:area_ulab_cifar10} and \Cref{fig:area_ulab_food101}. Again, we observe that \algo\ generates a more balanced distribution of pseudo-labels compared to FixMatch and FlexMatch, which is a key factor enabling our method to alleviate {aggregated biases}. Additionally, as depicted in \Cref{fig:line_conf_cifar10,fig:line_conf_food101}, the incorporation of DLS enables the model to more accurately assess the learning difficulty of various SSL tasks, thereby alleviating {cognitive deviation}. In contrast, the combination of VPT with FixMatch may lead to significant {cognitive deviation}, as shown in \Cref{fig:line_conf_cifar10}, leading to notable performance deterioration.

\vskip -0.2in

\subsection{Discussion on Cognitive Deviation and Calibration}
\label{app:cd}

\begin{figure*}[!h]
\begin{center}
   \begin{subfigure}[b]{0.24\textwidth}
   \centering
   \includegraphics[width=\linewidth,height=1.28in]{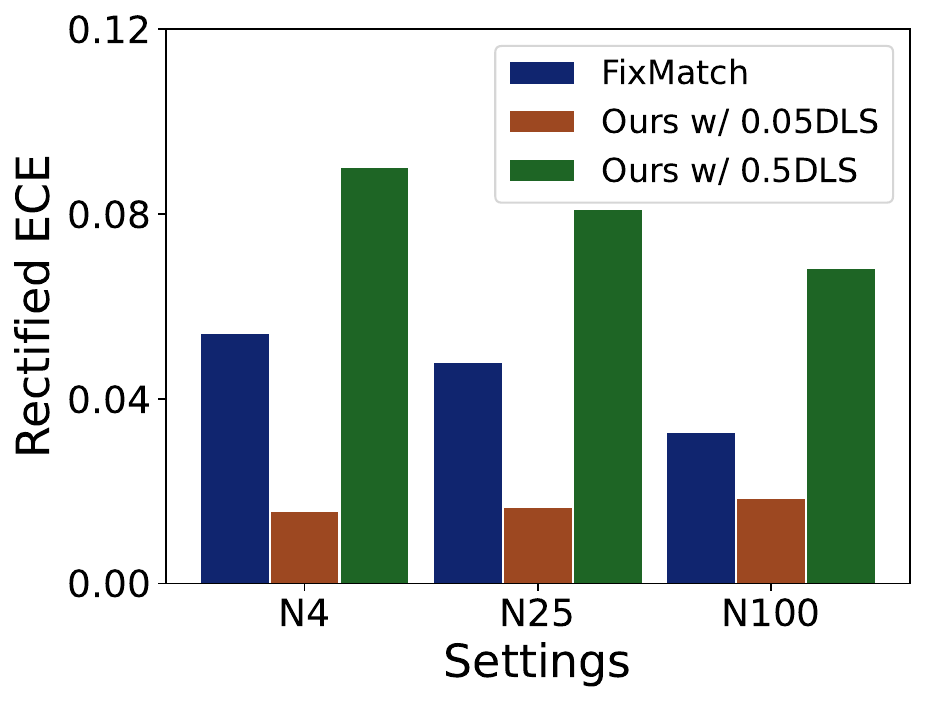}
   \vskip -0.06in

   \caption{}
   \label{fig:bar_ece}
   \end{subfigure}
   \begin{subfigure}[b]{0.24\textwidth}
   \centering
   \includegraphics[width=\linewidth,height=1.25in]{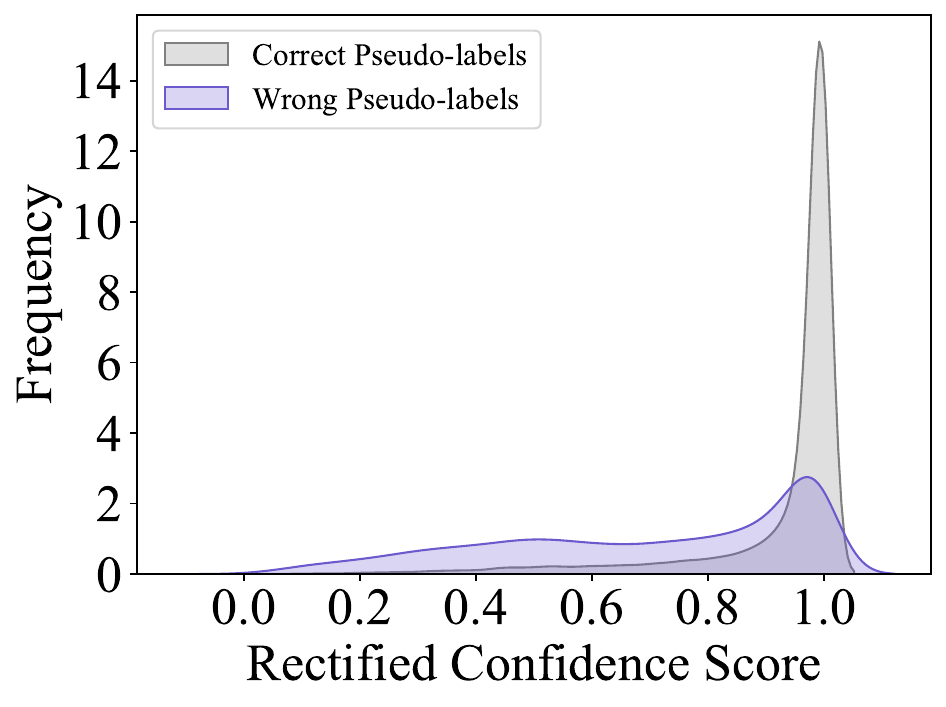}
   \vskip -0.06in
   \caption{}
   \label{fig:fix_conf}
   \end{subfigure}
   \begin{subfigure}[b]{0.24\textwidth}
   \centering
   \includegraphics[width=\linewidth,height=1.25in]{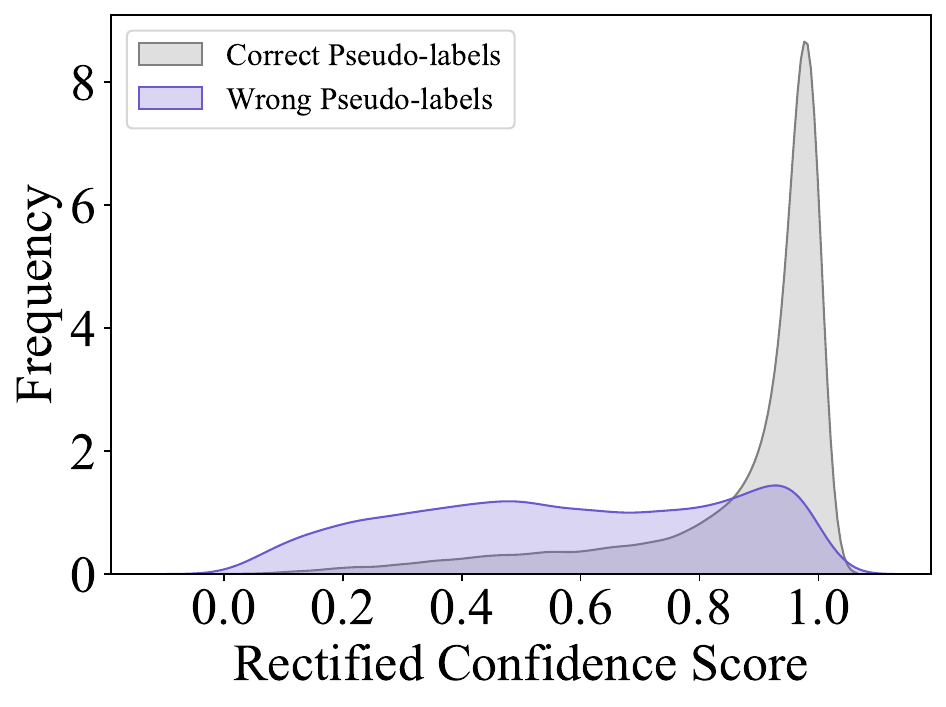}
   \vskip -0.06in

   \caption{}
   \label{fig:marlsm005_conf}
   \end{subfigure}
   \begin{subfigure}[b]{0.24\textwidth}
   \centering
   \includegraphics[width=\linewidth,height=1.25in]{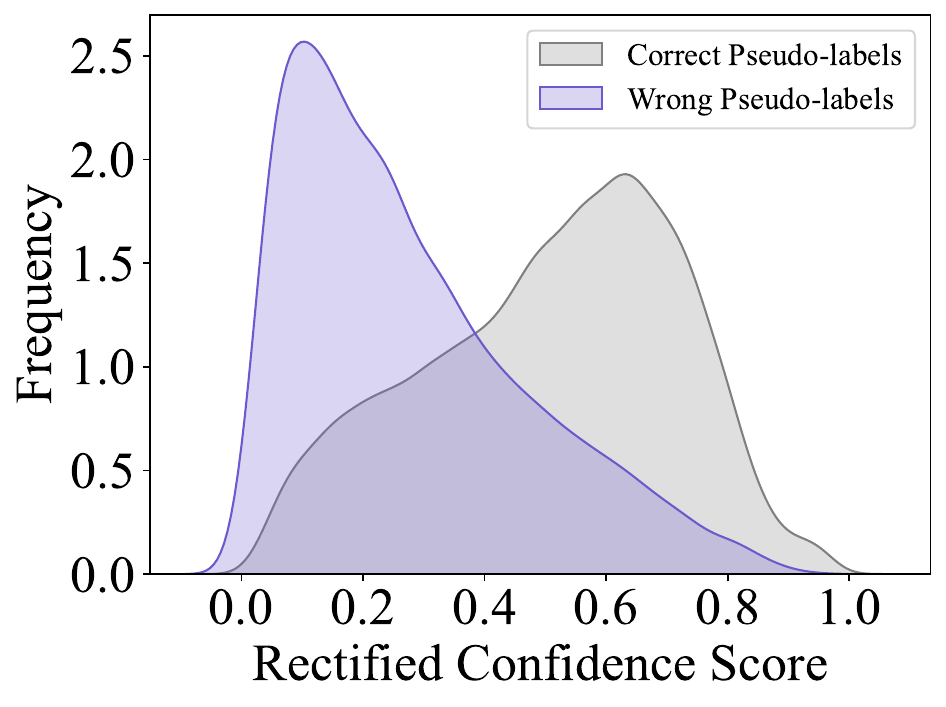}
   \vskip -0.06in

   \caption{}
   \label{fig:marlsm05_conf}
   \end{subfigure}
  \vskip -0.15in
\caption{(\labelcref{fig:bar_ece}): ECE for FixMatch and \algo\ on CIFAR-100. ``0.05DLS'' denotes $\lambda = 0.05$ in label smoothing, and ``0.5DLS'' denotes $\lambda = 0.5$. (\labelcref{fig:fix_conf,fig:marlsm005_conf,fig:marlsm05_conf}): Distribution of rectified confidences of FixMatch, \algo\ with $\lambda=0.05$ and $\lambda=0.5$, respectively.}
\label{fig:cd}
\end{center}
\vskip -0.15in
\end{figure*}

Model calibration \cite{guo2017calibration} is to modify the predictive probability which can reflect true correctness likelihood. However, cognitive deviation means the model does not exhibit corresponding confidence across settings of various learning difficulties. While there are indeed certain similarities between these two concepts, we show that mitigating cognitive deviation is better than improving calibration to boost performance.
First, we rectify the model output confidence to align with the accuracy of pseudo-labels. We denote $\kappa = \mathrm{ACC} / \frac{1}{\mu B} \sum_{j=1}^{\mu B} \max (\boldsymbol{q}_j)$, where ``ACC'' denotes the accuracy of pseudo-labels, then rectify confidences by $\kappa \cdot \max (\boldsymbol{q}_j)$.
Next, we show that the DLS proposed to overcome cognitive deviation can generate more reasonable sample weights for unlabeled data than models with better calibration.
\Cref{fig:bar_ece} shows the expected calibration error (ECE) of different models, \textit{i.e.,} FixMatch, \algo\ with $\lambda=0.05$ and $\lambda=0.5$ set in DLS. It can be seen that applying weak DLS ($\lambda=0.05$) improves calibration, but it hurts when applying strong DLS ($\lambda=0.5$). 
However, in \Cref{fig:fix_conf,fig:marlsm005_conf,fig:marlsm05_conf}, we observe that FixMatch exhibits significant overconfidence even in wrong predictions. Although applying weak DLS in \algo\ can partially alleviate this issue, the model assigns high weights for many unlabeled samples with wrong pseudo-labels. In contrast, strong DLS distinguishes correct and wrong pseudo-labels with a much clearer confidence boundary.
To sum up, we show that DLS does not improve the calibration, but helps correlate the model confidence with learning difficulties and generate conservative certainty for pseudo-labels.

\subsection{Why VPT is better than full fine-tuning?}
We will provide reasons from two perspectives. Firstly, a similar issue has been discussed in \cite{han2024facing}, which suggests that full fine-tuning can outperform VPT only when the pre-trained task and downstream task share similar objectives but have dissimilar data feature distributions. It is evident that SSL tasks fall outside the scope of this scenario, indicating that VPT is expected to perform better. Secondly, when considering full fine-tuning, the target tasks often require a substantial amount of data for effective adaptation \cite{dosovitskiy2020image,chen2022revisiting}. However, SSL tasks typically operate under the assumption of extremely limited labeled data availability. Therefore, VPT is considered more suitable for SSL tasks compared to full fine-tuning.

\section{Limitations of the work}

We conclude our limitations concisely below:
\vspace{-\topsep}
\begin{itemize}
\setlength\itemsep{0.1em}
\item Our work primarily focuses on the pre-trained CLIP model. However, it is important to note that there are other self-supervised pre-training methods available, such as SimCLR \cite{chen2020simple} and MOCO \cite{he2020momentum}. In future work, we plan to investigate the impact of these alternative pre-training methods on performance.
\item Our current approach only utilizes the image encoder in CLIP, neglecting the use of the text encoder. Considering the potential benefits of incorporating the text encoder, we believe it is worth exploring how its inclusion can augment the model's performance.
\end{itemize}


\end{document}